\documentclass[11pt,letterpaper]{article}
\usepackage[T1]{fontenc}
\usepackage[utf8]{inputenc}
\usepackage{lmodern}
\usepackage{amsmath,amssymb}
\usepackage[letterpaper,textwidth=6.65in,top=.87in,bottom=.87in,headheight=14pt,headsep=13pt,footskip=28pt]{geometry}
\usepackage{microtype,graphicx,booktabs,array,tabularx,longtable}
\usepackage{float,caption,placeins,fancyhdr,needspace}
\usepackage{xcolor}
\definecolor{linkink}{RGB}{31,64,91}
\usepackage[unicode,colorlinks=true,linkcolor=linkink,citecolor=linkink,urlcolor=linkink,bookmarksnumbered=true,bookmarksopen=false]{hyperref}
\hypersetup{pdftitle={Success Can Still Fail: Objective Choice in Recursive Self-Training},
  pdfauthor={Elijah Ng and Jennifer Dodgson},
  pdfsubject={Objective choice in recursive self-training},pdfdisplaydoctitle=true}
\input{glyphtounicode}
\title{\vspace{-3.5em}\LARGE\bfseries Success Can Still Fail:\\[.15em]
  Objective Choice in Recursive Self-Training}
\author{Elijah Ng\\[-.1em]{\small\href{mailto:elijah@eigenform.ai}{\texttt{elijah@eigenform.ai}}}
  \and Jennifer Dodgson\\[-.1em]{\small\href{mailto:jen@eigenform.ai}{\texttt{jen@eigenform.ai}}}}
\date{}
\begin{document}
\maketitle
\vspace{-2.5em}

\begin{abstract}

In recursive self-training, training exclusively on verified successes can still fail---yet the same retained data that yields a weak checkpoint under one objective can train a strong agent under another. We study this feedback loop on a file-cleanup task, fitting a fresh LoRA adapter on a frozen Qwen2.5-Coder-7B-Instruct base each round from up to three recent selected collections and a fixed rehearsal corpus. Across three starting datasets, we compare supervised fine-tuning (SFT) and probability-weighted Dynamic Fine-Tuning (DFT), each with and without a base-model forward-KL penalty. Both SFT variants halt between rounds 6 and 8 because they collect too few successful examples. DFT loses much of its early success, although two of its three runs reach round 10. Anchored Supervised Fine-Tuning (ASFT), which combines weighting and anchoring, completes all ten rounds in all three runs and reaches 92.3\% mean success on a fixed 200-item evaluation panel. Yet weak checkpoints coexist with useful buffers: DFT's round-6 checkpoints average 22.2\% success, while fresh ASFT fits on those same buffers reach 66.8\%, and mid-run switches from SFT or DFT to ASFT recover high late-round performance. In an idealized population model, we derive a sufficient condition for retained data distributions to move closer over a full buffer turnover. Empirically, however, round-3 DFT-to-ASFT switches nearly match ASFT throughout at round 9 (85.5\% vs. 85.8\%) even as their retained buffers grow farther apart in all three matched pairs. Recursive objectives should therefore be evaluated by how well their agents perform, whether they sustain collection, and what later agents can learn from their data.

\end{abstract}

\section{Introduction}\label{sec:1}

On a fixed dataset, a training objective is evaluated through the model it produces. In recursive self-training, that model also helps determine the next training dataset. A policy generates interactions, a selection rule retains successful records, and fitting turns those records into a new policy. Repetition couples the objective's immediate effect on the model to its later effects through the data. The relevant question is therefore whether a training objective supports continued learning across this feedback loop.

Learning from verified model outputs has established precedents. STaR trains on rationales that yield correct answers, and ReST-EM alternates generation, binary filtering, and supervised fitting on mathematics and coding tasks (\hyperlink{cite.zelikman2022}{Zelikman et al., 2022}; \hyperlink{cite.singh2024}{Singh et al., 2024}). ReST also compares alternative offline fitting losses within recursive training (\hyperlink{cite.gulcehre2023}{Gulcehre et al., 2023}). Our study examines objective choice in a protocol based on Negative Space Learning, which combines environment-mediated selection, fresh adapters on a shared pretrained base, rehearsal examples, and alternative data-retention rules (\hyperlink{cite.dodgson2026}{Dodgson et al., 2026}). Within this setting, we ask how training objectives change the course of a lineage and what its retained data can support when the resulting checkpoints perform poorly.

We study Qwen2.5-Coder-7B-Instruct on a file-cleanup task. An attempt succeeds when it removes at least 5\% of disposable files, preserves protected files, and leaves the system usable. Successful interaction records supply training rows. Each round fits a fresh LoRA adapter on the same frozen pretrained base using up to three recent selected collections and 500 fixed rehearsal examples. The lineage inherits its history through selected data: older behavior can persist beyond the explicit retention window when later agents reproduce it in new records. Within each run, all objectives share the initial selected collection, collection and selection procedures, and retention rule.

The comparison varies two components of existing supervised objectives: target-token weighting and reference regularization. SFT applies equally weighted target-token cross-entropy. Dynamic Fine-Tuning (DFT) weights each target's cross-entropy by the model's current probability of that token, with the weight detached during differentiation (\hyperlink{cite.wu2026}{Wu et al., 2026}). SFT+KL adds a forward-KL penalty toward the common pretrained base, while Anchored Supervised Fine-Tuning (ASFT) combines this penalty with DFT's probability weighting (\hyperlink{cite.zhu2026}{Zhu et al., 2026}). We follow these four objectives across three distinct starting datasets with a common training seed, measuring online collection success and performance on a fixed 200-item evaluation panel.

The objectives produce sharply different trajectories. ASFT completes all ten rounds in every run and reaches 92.3\% mean fixed-panel success at round 10. SFT and SFT+KL stop between rounds 6 and 8 because they collect too few selected rows to continue. DFT improves rapidly at first but loses much of that success. At round 6, the last checkpoint available for every objective in every run, ASFT achieves 61.5\% mean evaluation success, compared with 22.2\% for DFT, 3.2\% for SFT, and 3.0\% for SFT+KL. In this setting, we find that neither probability weight nor reference regularization alone can account for the observed performance over the rounds.

The same retained data that produced a weak DFT checkpoint supported a much stronger checkpoint when fitted with ASFT. We test this by fitting alternative objectives to the recorded round-6 buffers. DFT-generated buffers yield 66.8\% mean evaluation success under fresh ASFT fits, compared with 22.2\% for their original DFT fits. Data history still matters: the same training objective performs differently across source buffers. We also change objectives during ongoing recursion, retaining the existing data while fitting each new adapter from the common base. SFT and DFT beginnings remain compatible with high later success after a change to ASFT. These experiments distinguish the current agent's performance, its buffer's usefulness for another fit, and its ability to collect the next successful corpus.

Finally, we formalize the ordered retained buffer as the state of a recursive feedback process, drawing on stateful performative prediction (\hyperlink{cite.brown2022}{Brown et al., 2022}). Under fixed mixture weights and Lipschitz fitting and collection maps, we give a sufficient condition for contraction over a full retention-window turnover. This conditional result separates the sensitivities of fitting and selected-data collection, while assigning rehearsal mixing and base-model regularization distinct roles. Its assumptions are not established for the implemented protocol, and convergence to a self-consistent model-data pair does not imply task success. Together, the experiments and formalization show why recursive self-training objectives should be assessed through both the data passed forward and the checkpoints they produce.

\FloatBarrier

\section{Related work}\label{sec:2}

\subsection{Verified self-training and data-mediated inheritance}\label{sec:2.1}

Verified self-training combines data collection, selection, and model fitting. ReST separates generation from offline fitting and compares several losses for fine-tuning the previous policy (\hyperlink{cite.gulcehre2023}{Gulcehre et al., 2023}). RAFT selects responses by learned reward ranking, with distillation experiments that use different models for generation and fitting (\hyperlink{cite.dong2023}{Dong et al., 2023}). Our file-cleanup task instead selects training data through execution-based success checks. Correctness filtering has precedents in reasoning: STaR retains rationales yielding correct answers, and ReST-EM filters mathematical answers or code using task-specific checks (\hyperlink{cite.zelikman2022}{Zelikman et al., 2022}; \hyperlink{cite.singh2024}{Singh et al., 2024}). Both methods refit the original pretrained model each round, demonstrating that information can pass between rounds through selected data without retaining the preceding round's fitted weights.

Our protocol builds on Negative Space Learning (NSL), which combines environment-mediated selection, fresh adapters on a shared pretrained base, rehearsal data, and alternative retention policies, including windows of recent collections (\hyperlink{cite.dodgson2026}{Dodgson et al., 2026}). In our Qwen2.5-Coder-7B-Instruct file-cleanup experiments, each round fits a fresh adapter on the same pretrained base using up to three recent collections and a fixed set of rehearsal examples. The resulting corpus at each round is hence a mix of synthetic and rehearsal data; collection is controlled by a row target rather than a total-token target. It is within this setting that we explore the dynamics of retained data buffers and the selection of training objectives. This includes what the retained data can support under other objectives when the trained checkpoint otherwise performs poorly.

\subsection{Training objectives: probability weighting and reference regularization}\label{sec:2.2}

\hyperlink{cite.qin2025}{Qin and Springenberg (2025)} interpret supervised fine-tuning (SFT) on success-filtered data as maximizing a lower bound on a sparse-reward reinforcement-learning objective. Our study compares four existing supervised training objectives distinguished by target-token weighting and fixed-base reference regularization. SFT uses cross-entropy with equal weights on target tokens. Dynamic Fine-Tuning (DFT) multiplies each token's cross-entropy by the model's current probability of that target, treating the multiplier as constant during differentiation (\hyperlink{cite.wu2026}{Wu et al., 2026}). This reduces the contribution of low-probability targets relative to SFT. SFT+KL adds a forward KL divergence term, KL(base \ensuremath{\Vert} current), as a soft penalty toward the common pretrained base rather than the preceding round's model. Anchored Supervised Fine-Tuning (ASFT) combines this penalty with DFT (\hyperlink{cite.zhu2026}{Zhu et al., 2026}).

Within recursive training, ReST compares offline losses, and \emph{Progress or Regress?} studies iterative SFT, preference optimization, and their alternation (\hyperlink{cite.gulcehre2023}{Gulcehre et al., 2023}; \hyperlink{cite.wu2025}{Wu et al., 2025}). ForTIFAI addresses recursive degradation by suppressing high-confidence target updates (\hyperlink{cite.zibakhshshabgahi2026}{Zibakhsh Shabgahi et al., 2026}), whereas DFT reduces the contributions of low-probability targets relative to SFT.

\subsection{Recursive data dynamics: curation, retention, and stateful feedback}\label{sec:2.3}

Data retention and curation shape the distributions learned over successive training rounds. \hyperlink{cite.shumailov2024}{Shumailov et al. (2024)} characterize model collapse through loss of fidelity to the original distribution, beginning with its tails. \hyperlink{cite.gerstgrasser2024}{Gerstgrasser et al. (2024)} establish that retaining the original dataset and accumulating all subsequent synthetic datasets keeps expected prediction error bounded in their linear-regression model. Their empirical comparisons with data replacement include controls matched for dataset growth. Our protocol instead retains up to three recent collections alongside rehearsal data. \hyperlink{cite.ferbach2024}{Ferbach et al. (2024)} show that curation can implicitly optimize preferences: under idealized fitting, probability can concentrate on preferred outputs while diversity declines.

Retention makes recursive training stateful. \hyperlink{cite.brown2022}{Brown et al. (2022)} formalize model-dependent distribution changes with persistent state, and \hyperlink{cite.zhao2025}{Zhao et al. (2025, v3 2026)} derive curated population updates with conditional contraction under reference mixing. While mixing reference examples into the training distribution is distinct from adding a base-model KL penalty, we use this to describe the feedback structure. We also use much of the framework established by Brown to describe conditional stability in our recursive training protocol.

\FloatBarrier

\section{Self-Training Approach and Objective Functions}\label{sec:3}

\subsection{Recursive protocol}\label{sec:3.1}

We study a recursive collect--select--fit protocol based on Negative Space Learning (\hyperlink{cite.dodgson2026}{Dodgson et al., 2026}). An agent interacts with an environment, and a fixed task-specific success criterion selects the interaction records used for training. Each round fits a fresh LoRA adapter on the same frozen pretrained base. We call a sequence of rounds linked through selected training data a \emph{self-training lineage}.

Let $\pi_0$ be the pretrained base policy and $A_0$ an initial selected collection generated by that policy. Within each run, all objectives share $A_0$. Before the first fit, the retained buffer is $B_1=(A_0)$. At model round $t\geq1$, the fitting corpus $D_t$ combines the collections in $B_t$ with a fixed rehearsal corpus $R$. With training objective $u_t$, one round consists of

\[
\begin{aligned}
\pi_t &= \operatorname{Fit}_{u_t}(\pi_0,D_t),\\
A_t &= \operatorname{CollectSelect}(\pi_t),\\
B_{t+1} &= \operatorname{Window}_K(B_t,A_t).
\end{aligned}
\]

Here $\operatorname{Fit}$ trains a new adapter from the common base. $\operatorname{CollectSelect}$ samples agent interactions, retains successful attempts, and converts their records into training rows. The task-environment distribution, success criterion, and record-conversion procedure are held fixed. The ordered buffer retains the $K$ most recent collections, with the newest collection first. In our experiments, $K=3$. Each collection targets 1,000 selected rows, and each fit includes 500 rehearsal rows. We use a 16k-token maximum training sequence length; Negative Space Learning uses 4,500 rows per collection and reports a 1k-token limit (\hyperlink{cite.dodgson2026}{Dodgson et al., 2026, \S{}5.2 and Annex II}). A successful attempt can supply several training rows. A lineage stops if collection yields too few selected rows to continue.

\begin{figure}[htbp]
\centering
\includegraphics[width=\linewidth,height=.74\textheight,keepaspectratio]{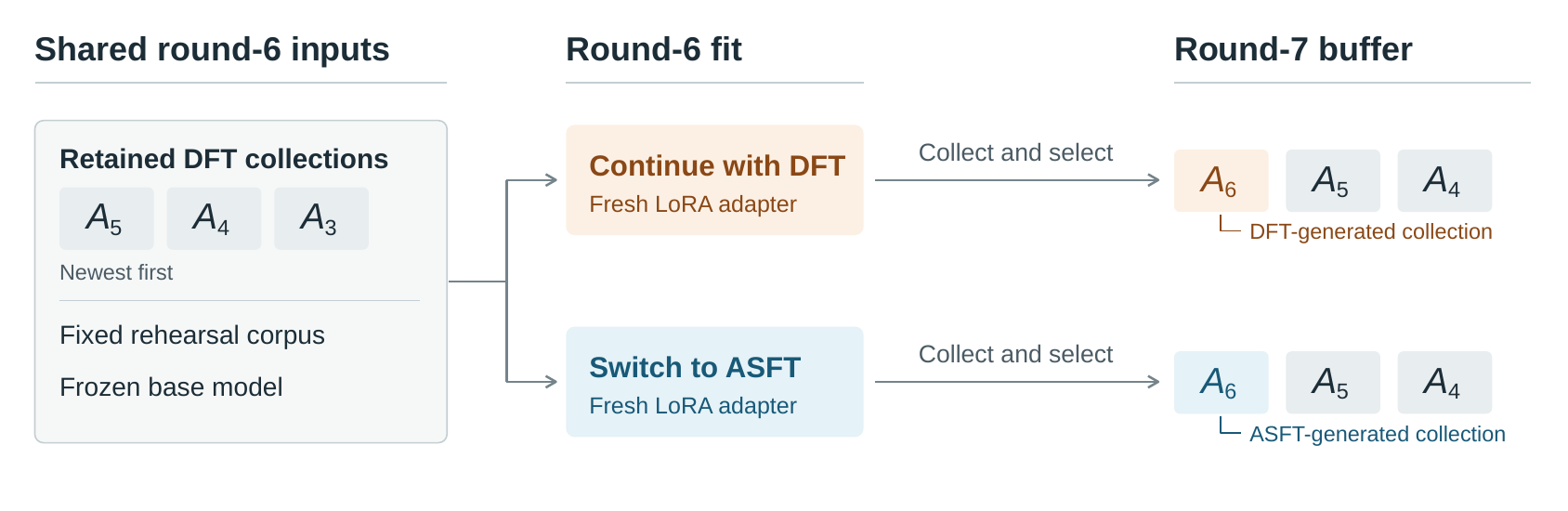}
\caption[One objective switch across two fits]{\textbf{One objective switch across two fits.} At round 6, both branches fit fresh adapters from the same frozen base using the same three DFT-generated collections and fixed rehearsal corpus. Each agent then generates its own collection $A_6$ from successful attempts. At round 7, the two buffers share the inherited collections $A_5$ and $A_4$.}
\label{fig:1}
\end{figure}

Information passes between rounds through selected data. The window limits direct retention, but behavior from older rounds can persist when later agents reproduce it in newly selected records. A constant-objective lineage fixes $u_t=u$. An objective switch changes the training objective while retaining the existing buffer, so the new objective initially acts on data produced under the preceding policy. \hyperref[fig:1]{Figure~1} illustrates a switch from DFT to ASFT at model round 6.

\subsection{Training objectives}\label{sec:3.2}

We use \emph{training objective} for the loss and its differentiation convention, and \emph{fitting procedure} for the complete process used to train an adapter. In the population analysis, the corresponding \emph{fitting map} $F_u$ returns the policy produced from a training distribution under objective $u$, with the common base and remaining fitting settings fixed.

We compare four objectives organized by target-token weighting and reference regularization. Let $\pi_\theta$ denote the policy during adapter fitting. At each valid target-token position $i$ in $D_t$, let $h_i$ be the recorded prefix, $y_i$ the target token, and $p_i=\pi_\theta(y_i\mid h_i)$. With $N_t$ valid target positions across both selected and rehearsal examples, the computed training loss is

\[
\mathcal{L}_u(\theta;D_t)
=-\frac{1}{N_t}\sum_i w_u(p_i)\log p_i
+\lambda_u\mathcal{K}(\theta;D_t),
\]

where

\[
\mathcal{K}(\theta;D_t)
=\frac{1}{N_t}\sum_i
D_{\mathrm{KL}}\!\left(
\pi_0(\cdot\mid h_i)\,\Vert\,\pi_\theta(\cdot\mid h_i)
\right).
\]

\begin{center}
\small
\renewcommand{\arraystretch}{1.17}
\begin{tabularx}{\linewidth}{@{}>{\raggedright\arraybackslash}X>{\raggedright\arraybackslash}X>{\raggedright\arraybackslash}X@{}}
\toprule
\textbf{Target-token weighting} & \textbf{No reference penalty:}\newline $\lambda_u=0$ & \textbf{Base reference penalty:}\newline $\lambda_u>0$ \\
\midrule
Equal weights:\newline $w_u(p)=1$ & SFT & SFT+KL \\
Probability weights:\newline $w_u(p)=\operatorname{sg}(p)$ & DFT & ASFT \\
\bottomrule
\end{tabularx}
\end{center}

The stop-gradient operator $\operatorname{sg}$ preserves the weight in the forward pass and treats it as a constant during differentiation. DFT (\hyperlink{cite.wu2026}{Wu et al., 2026}) recomputes these weights from the current model, reducing the contribution of low-probability targets relative to SFT. SFT+KL adds a forward KL penalty toward the fixed base policy; ASFT (\hyperlink{cite.zhu2026}{Zhu et al., 2026}) combines that penalty with probability weighting. This KL term compares the base and fitted models' full next-token distributions at each valid target-token position in the recorded training buffer, conditioned on the same set of preceding tokens from that example. We set the forward-KL coefficient to $\lambda_u=0.1$ for both SFT+KL and ASFT, following \hyperlink{cite.zhu2026}{Zhu et al. (2026, \S{}5.3, Fig. 2)}, whose coefficient sweep on medical QA benchmarks favored values near 0.1.

For DFT and ASFT, the detached weighting also admits a useful gradient-equivalent potential. At each target position,

\[
\nabla_\theta[-\operatorname{sg}(p_i)\log p_i]
=-p_i\nabla_\theta\log p_i
=-\nabla_\theta p_i.
\]

Consequently, ASFT's training gradient equals the gradient of

\[
\widetilde{\mathcal{L}}_{\mathrm{ASFT}}(\theta;D_t)
=-\frac{1}{N_t}\sum_i p_i
+\lambda_{\mathrm{ASFT}}\mathcal{K}(\theta;D_t).
\]

\begin{figure}[htbp]
\centering
\includegraphics[width=\linewidth,height=.74\textheight,keepaspectratio]{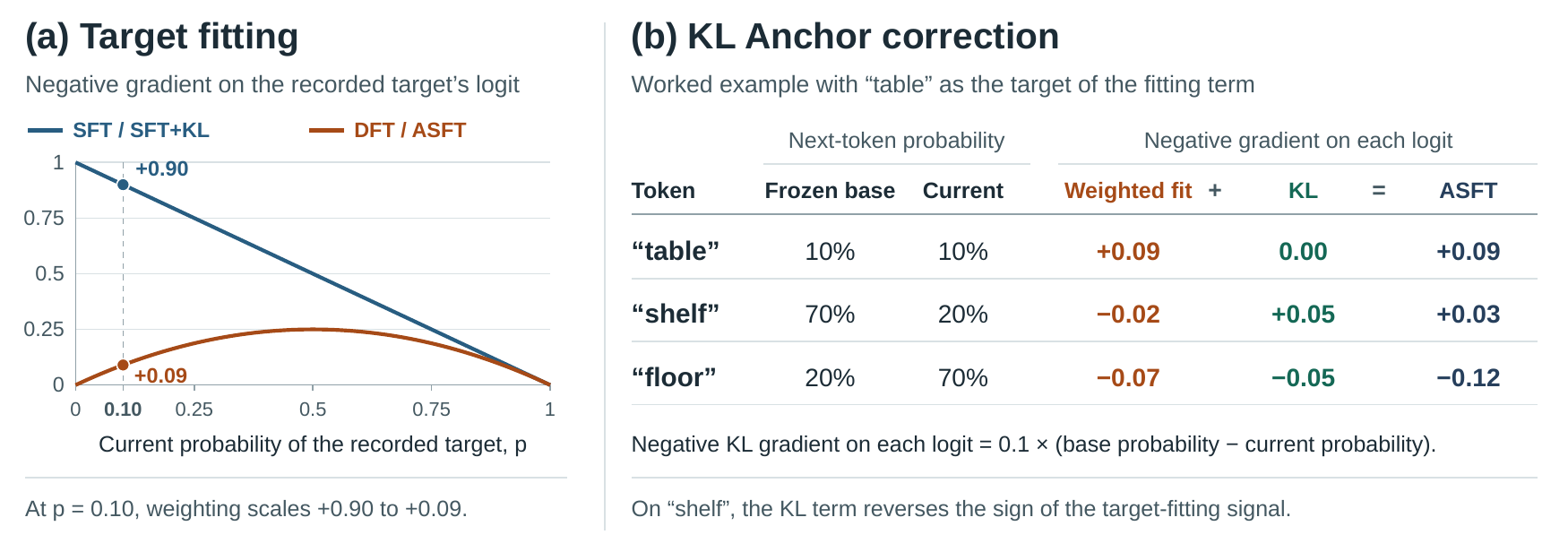}
\caption[Target weighting and KL anchoring at one recorded prefix]{\textbf{Target weighting and KL anchoring at one recorded prefix.} (a) With current target probability $p$, the target-fitting contribution to the negative gradient on the recorded target's logit is $1-p$ under equal weighting and $p(1-p)$ under detached probability weighting. (b) For each token $v$ at prefix $h$, ASFT adds the KL contribution $\lambda_u[\pi_0(v\mid h)-\pi_\theta(v\mid h)]$ to the weighted target-fitting contribution. On the ``shelf'' logit, the two contributions sum to $-0.02+0.05=+0.03$.}
\label{fig:2}
\end{figure}

The scalar potential yields the same training gradient as the detached objective. \hyperref[sec:A.3]{Appendix~A.3} derives the curvature of this potential at a fixed prefix in probability coordinates. \hyperref[fig:2]{Figure~2} illustrates how the target-fitting and KL terms combine in logit space at one recorded prefix.

\subsection{Stateful feedback and conditional stability}\label{sec:3.3}

The state of the recursive self-training process is encapsulated by its retained data buffer. A model is fitted to that state, generates a new selected collection, and thereby changes the state used by later fits. \hyperlink{cite.brown2022}{Brown et al. (2022)} studied coupled distribution--model updates and characterized stable pairs through mutual consistency of the distribution and its fitted model. Their finite-memory case is analogous to what we characterize here as a sliding window. We hence use it to formulate a conditional stability result for our protocol.

Replace finite collections by distributions over valid target-token instances $z=(h,y)$. Once the window is full, write its ordered state as $\mathbf{B}_t=(Q_{t-1},\ldots,Q_{t-K})$. Let $Q_R$ be the distribution of the fixed rehearsal corpus. Consider the idealized fitting distribution

\[
P(\mathbf{B}_t)
=\rho Q_R+(1-\rho)\sum_{j=1}^{K}a_jQ_{t-j},
\qquad a_j\geq0,\quad\sum_{j=1}^{K}a_j=1.
\]

Here the rehearsal mass $0\leq\rho<1$ and age weights $a_j$ are fixed. Consider the fitting and collection steps along either branch of \hyperref[fig:1]{Fig.~1}. In the population model, the deterministic fitting map $F_u$ takes a training distribution $P$ and returns a policy $F_u(P)$ under objective $u$, with the common base and remaining fitting settings fixed. The collection map $C$ takes a policy $\pi$ and returns the distribution $C(\pi)$ of selected training tokens under the fixed environment and selection procedure. Define

\[
\begin{aligned}
Q_t &= C\!\left(F_u(P(\mathbf{B}_t))\right),\\
\Phi_u(\mathbf{B}_t)&=(Q_t,Q_{t-1},\ldots,Q_{t-K+1}).
\end{aligned}
\]

The operational protocol uses finite samples and finite-compute fits. The population system isolates its feedback structure. Under token-normalized training, the realized mixture weights depend on the token totals of the selected collections and rehearsal corpus.

\phantomsection\label{prop:1}\noindent
\textbf{Proposition 1 --- Conditional stability with a finite retention window.} Equip token distributions with Wasserstein distance $W_1$ under a fixed ground metric, policies with a metric $d_\pi$, and ordered buffers with

\[
d_B(\mathbf{B},\mathbf{B}')=\max_{1\leq j\leq K}W_1(Q_j,Q_j').
\]

Assume the maps are defined on a nonempty complete invariant buffer domain and satisfy, throughout the corresponding input domains,

\[
\begin{aligned}
d_\pi(F_u(P),F_u(P'))&\leq L_{\mathrm{fit},u}W_1(P,P'),\\
W_1(C(\pi),C(\pi'))&\leq L_{\mathrm{collect}}d_\pi(\pi,\pi').
\end{aligned}
\]

If

\[
q_u=(1-\rho)L_{\mathrm{fit},u}L_{\mathrm{collect}}<1,
\]

then the $K$-round map is contractive:

\[
d_B(\Phi_u^K(\mathbf{B}),\Phi_u^K(\mathbf{B}'))
\leq q_u\,d_B(\mathbf{B},\mathbf{B}').
\]

It follows that the system has a unique fixed buffer $\mathbf{B}_u^*=(Q_u^*,\ldots,Q_u^*)$ and associated fitted policy $\pi_u^*=F_u(\rho Q_R+(1-\rho)Q_u^*)$, with $C(\pi_u^*)=Q_u^*$. Starting at any admissible full buffer,

\[
d_B(\mathbf{B}_t,\mathbf{B}_u^*)
\leq q_u^{\lfloor(t-t_0)/K\rfloor}
d_B(\mathbf{B}_{t_0},\mathbf{B}_u^*),\qquad t\geq t_0.
\]

The condition separates two sensitivities: how much the fitted policy changes when its training distribution changes, and how much the distribution of newly selected training data changes when that policy changes. \hyperref[sec:A.3]{Appendix~A.3} describes how the choice of training objective shapes the fitting step at the level of an individual target; for ASFT, probability weighting reduces the target-fitting term for low-probability targets, giving the base-model anchor more relative weight. In the collection step, a cleanup strategy can become more common in the next training collection if the policy uses it more often on successful attempts. We examine both sides of this loop in \hyperref[sec:4.3]{\S{}4.3} by tracking the fitted policy's predictions on fixed contexts and using the base model to score the newest selected targets.

At each round, one new collection enters the buffer, while the older remaining collections are carried forward unchanged. As the distance between two buffer states is defined as the maximum Wasserstein distance between collections in matching positions, if a pair attaining this maximum is carried forward unchanged, the distance between the buffers remains unchanged for that round. The contraction bound therefore applies every $K$ rounds, after every collection in the starting buffers has been replaced (the resulting distance after such an interval is also at most $q_u$ times its initial value). Assuming that $q_u < 1$, a longer window means a slower convergence rate in the bound. The condition $q_u<1$ itself does not depend on the window length. Proof for this is in \hyperref[sec:A.1]{Appendix~A.1}.

\begin{figure}[htbp]
\centering
\includegraphics[width=\linewidth,height=.74\textheight,keepaspectratio]{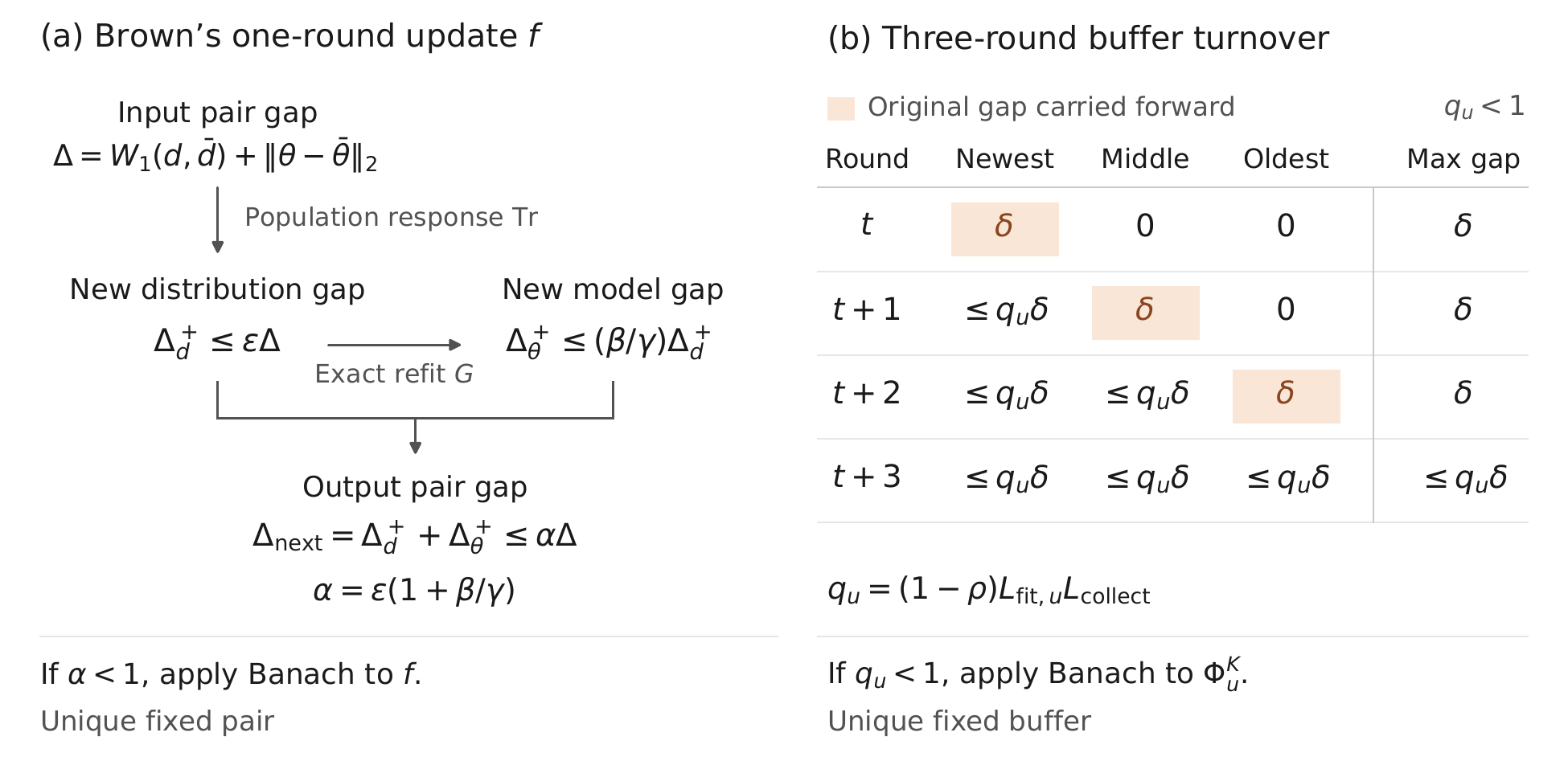}
\caption[Contraction after one update and after a full buffer turnover]{\textbf{Contraction after one update and after a full buffer turnover.} (a) In \hyperlink{cite.brown2022}{Brown et al. (2022, Theorem 4)}, the population response and exact refit together reduce the distance between distribution--model pairs by a factor of at most $\alpha$ when $\alpha<1$. (b) In the three-collection example from \hyperref[prop:1]{Proposition~1}, the initial gap $\delta$ starts in the newest slot and is carried forward unchanged through the next two updates (shaded cells). When that collection leaves the buffer, the maximum gap is at most $q_u\delta$.}
\label{fig:3}
\end{figure}

\hyperref[sec:4.4]{\S{}4.4} reports task and collection success after objective switches. \hyperref[sec:4.5]{\S{}4.5} measures how paired buffer distances change over complete turnovers under ASFT.

\FloatBarrier

\section{Experiments and Analysis}\label{sec:4}

We evaluate how training objectives affect task performance, continued collection, and the usefulness of retained data during recursive training. We study recursive training of Qwen2.5-Coder-7B-Instruct on a file-cleanup task. An attempt succeeds when the agent removes at least 5\% of the disposable files, leaves protected files intact, and keeps the system usable. Successful attempts supply the selected data for later rounds. Each round fits a fresh adapter from the common pretrained base using up to three recent collections and 500 fixed rehearsal examples. Each collection targets 1,000 selected rows.

We compare four objectives, SFT, DFT, SFT+KL, and ASFT, across three runs, each beginning with a distinct selected collection generated by the base policy. Within each run, all objectives share that starting collection. All runs use the same training seed. Online success is the fraction of collection episodes that succeed. Fixed-panel success measures performance on the same 200 evaluation items spanning five task variations. We report equal-weighted run means and descriptive minimum--maximum ranges.

\subsection{ASFT sustains successful recursion across ten rounds}\label{sec:4.1}

\begin{figure}[htbp]
\centering
\includegraphics[width=\linewidth,height=.74\textheight,keepaspectratio]{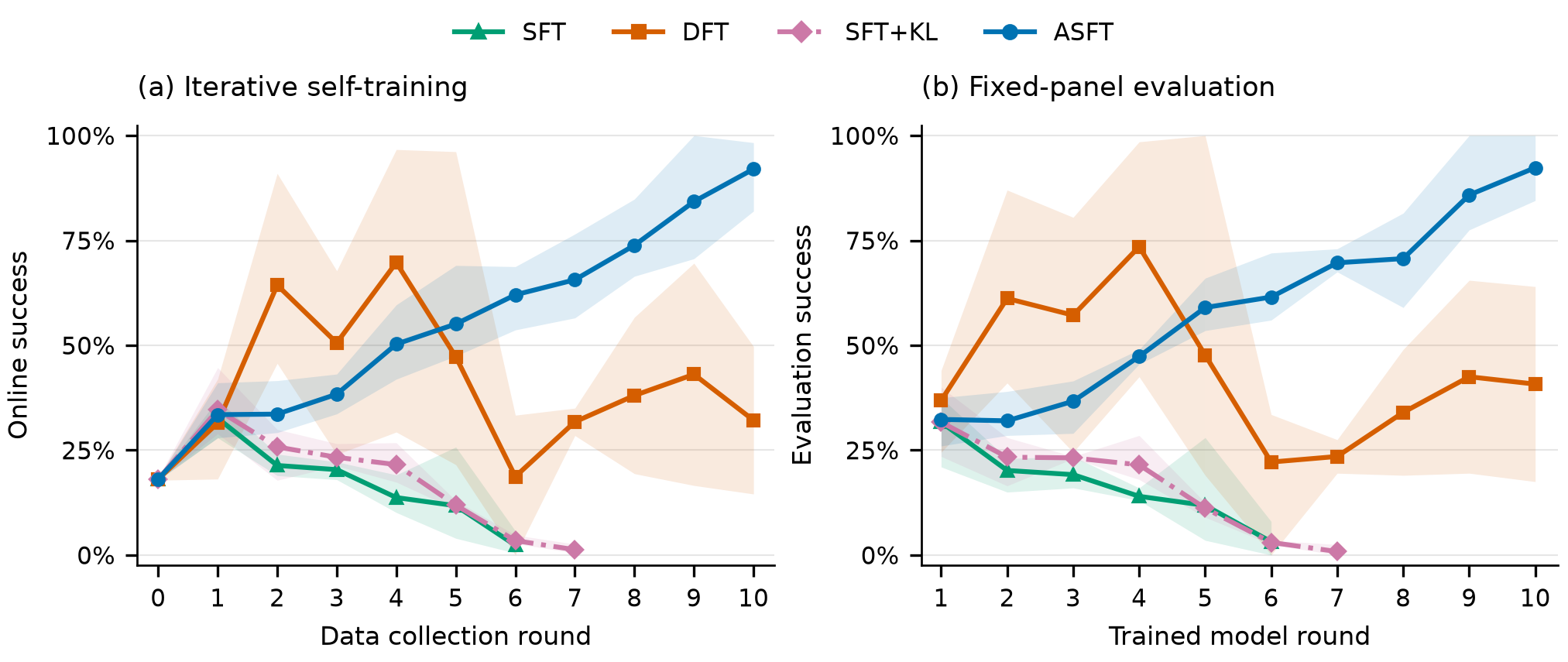}
\caption[ASFT sustains iterative self-training]{\textbf{ASFT sustains iterative self-training.} (a) Online success during data collection. Round 0 is the shared starting collection using the base model. (b) Evaluation success on the fixed 200-item panel across training rounds. For Run 1, the round-6 point uses the dedicated reevaluation reported in \hyperref[fig:5]{Fig.~5a}. Its other rounds use the August sweep. Points show equally weighted means across contributing runs. Shaded bands span their observed minima and maxima. \hyperref[fig:A1]{Fig.~A1} shows the individual runs. SFT and SFT+KL end when successful data no longer replenish the training buffer.}
\label{fig:4}
\end{figure}

The objectives produce sharply different recursive outcomes (\hyperref[fig:4]{Fig.~4a}). ASFT continues through all ten rounds in every run, reaching 92.0\% mean online success at round 10. Both SFT and SFT+KL stop between rounds 6 and 8 because they no longer collect enough successful data to replenish the training buffer. DFT improves rapidly at first but does not maintain its early success: one run stops at round 6, while two continue through round 10 with 32.1\% mean online success among those continuing runs.

ASFT's advantage also appears on the fixed evaluation panel (\hyperref[fig:4]{Fig.~4b}). At round 6, the last checkpoint available for all four objectives in all three runs, dedicated evaluations of the original checkpoints yield mean success rates of 61.5\% for ASFT, 22.2\% for DFT, 3.2\% for SFT, and 3.0\% for SFT+KL. ASFT leads in each of the three runs. At this checkpoint, ASFT outperforms both probability weighting alone and anchoring alone.

At round 10, ASFT achieves 92.3\% mean fixed-panel success across all three runs, with a range of 84.5--100\%. DFT achieves 40.8\% among its two continuing runs, which score 64.0\% and 17.5\%. SFT and SFT+KL decline toward zero before their lineages stop: their final checkpoints score 2/200 for Run 3 SFT at round 7, 3/200 for Run 2 SFT+KL at round 8, and 0/200 for Run 3 SFT+KL at round 7 (Appendix \hyperref[fig:A1]{Fig.~A1}).

These SFT outcomes differ from the improvements reported in Negative Space Learning, although our experiments do not isolate the effects of the different collection targets and maximum training sequence lengths relative to that setup. Our complementary single-round FictionalQA study in \hyperref[sec:A.9]{Appendix~A.9} also suggests that SFT might be preferable under different settings and that the relative performance of the training objectives could depend on the task. Even within the cleanup task, the ranking of objectives changed across rounds. DFT led in mean online success in collection rounds 2--4, and ASFT took the lead in round 5 (\hyperref[fig:4]{Fig.~4a}).

Our broader question is how the choice of training objective shapes a lineage across rounds, both in the agents it produces and in the data it carries forward. We next ask whether the same buffers used to train weaker checkpoints can support more successful agents under a different training objective.

\subsection{Weak checkpoints can coexist with useful retained data}\label{sec:4.2}

\begin{figure}[htbp]
\centering
\includegraphics[width=\linewidth,height=.74\textheight,keepaspectratio]{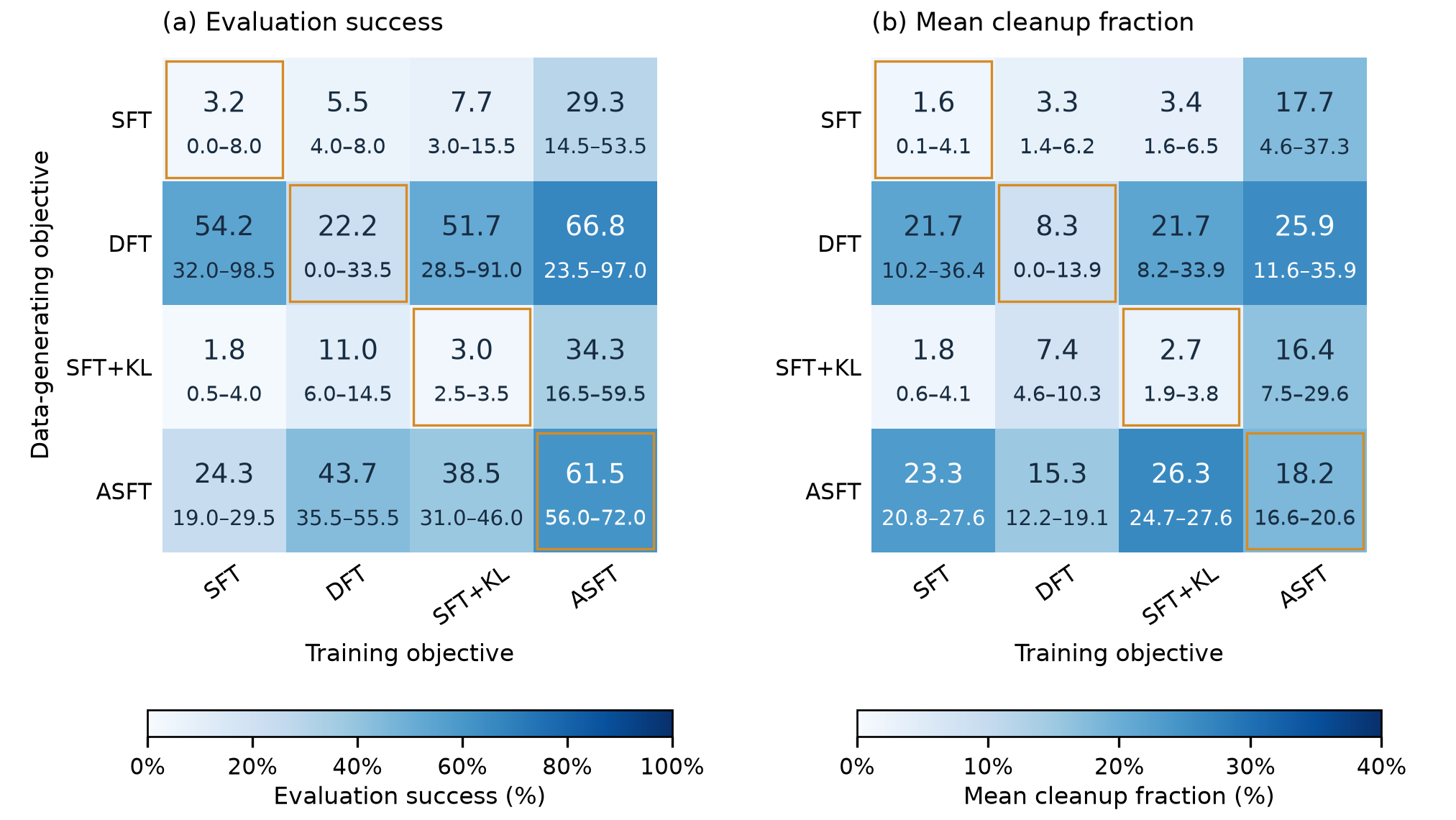}
\caption[Data from poorly performing lineages can still train successful agents]{\textbf{Data from poorly performing lineages can still train successful agents.} Rows identify the objective used to train the agents that collected the data; columns identify the objective used to train on those data. Each lineage supplies its recorded round-6 training buffer. Outlined diagonals reuse the original self-training fits; other cells train new adapters from the same base under the changed training objective. (a) Evaluation success and (b) mean fraction of disposable files cleared across all 200 items, including failures. Cells show equally weighted means across three lineages, with minimum--maximum ranges below.}
\label{fig:5}
\end{figure}

To separate checkpoint performance from the usefulness of retained data, we cross each recorded round-6 training buffer with all four training objectives (\hyperref[fig:5]{Fig.~5}). Rows identify the data-generating lineage's objective; columns identify the training objective. Diagonal cells reuse the original round-6 fits, while off-diagonal cells fit fresh adapters from the common base to the same recorded buffers. Reading across a row compares realized fits of the same data; reading down a column compares different data histories under a common training objective.

DFT provides the clearest separation. Its original round-6 fits achieve 22.2\% mean evaluation success, whereas fresh fits of those buffers achieve 54.2\% with SFT, 51.7\% with SFT+KL, and 66.8\% with ASFT. The retained collection therefore supports behaviour that is substantially more successful than the original checkpoint expresses.

Evaluation success depends on both the source of the data and the fitting objective. ASFT achieves the highest mean success for every data source in \hyperref[fig:5]{Fig.~5a}, but performs substantially better on DFT- and ASFT-generated buffers than on SFT- and SFT+KL-generated buffers. Holding ASFT fixed, mean success ranges from 29.3\% on SFT-generated buffers to 66.8\% on DFT-generated buffers. Together with the gains from refitting DFT's buffers, these differences show how inherited data and the current training objective jointly shape performance. \hyperref[sec:A.8]{Appendix~A.8} reports training-time agreement measurements for the ASFT fits.

Success and cleanup thoroughness also differ (\hyperref[fig:5]{Fig.~5b}). On ASFT-generated data, ASFT achieves 61.5\% success and clears 18.2\% of disposable files on average in the evaluation setting. SFT+KL clears 26.3\% but succeeds on 38.5\% of items. These clearance averages include failures. Clearance measures the fraction of disposable files removed; success requires both crossing the cleanup threshold and satisfying the safety/functionality checks.

Continued recursion required enough selected rows for the next fit (\hyperref[fig:4]{Fig.~4a}). Yet, we observe that collected data buffers can sometimes support a stronger checkpoint when fit under a different training objective (\hyperref[fig:5]{Fig.~5a}). A checkpoint evaluation score therefore only provides a partial assessment of a lineage's capacity for continued training.

\FloatBarrier

\subsection{Recent and historical targets across objectives}\label{sec:4.3}

\begin{figure}[htbp]
\centering
\includegraphics[width=\linewidth,height=.74\textheight,keepaspectratio]{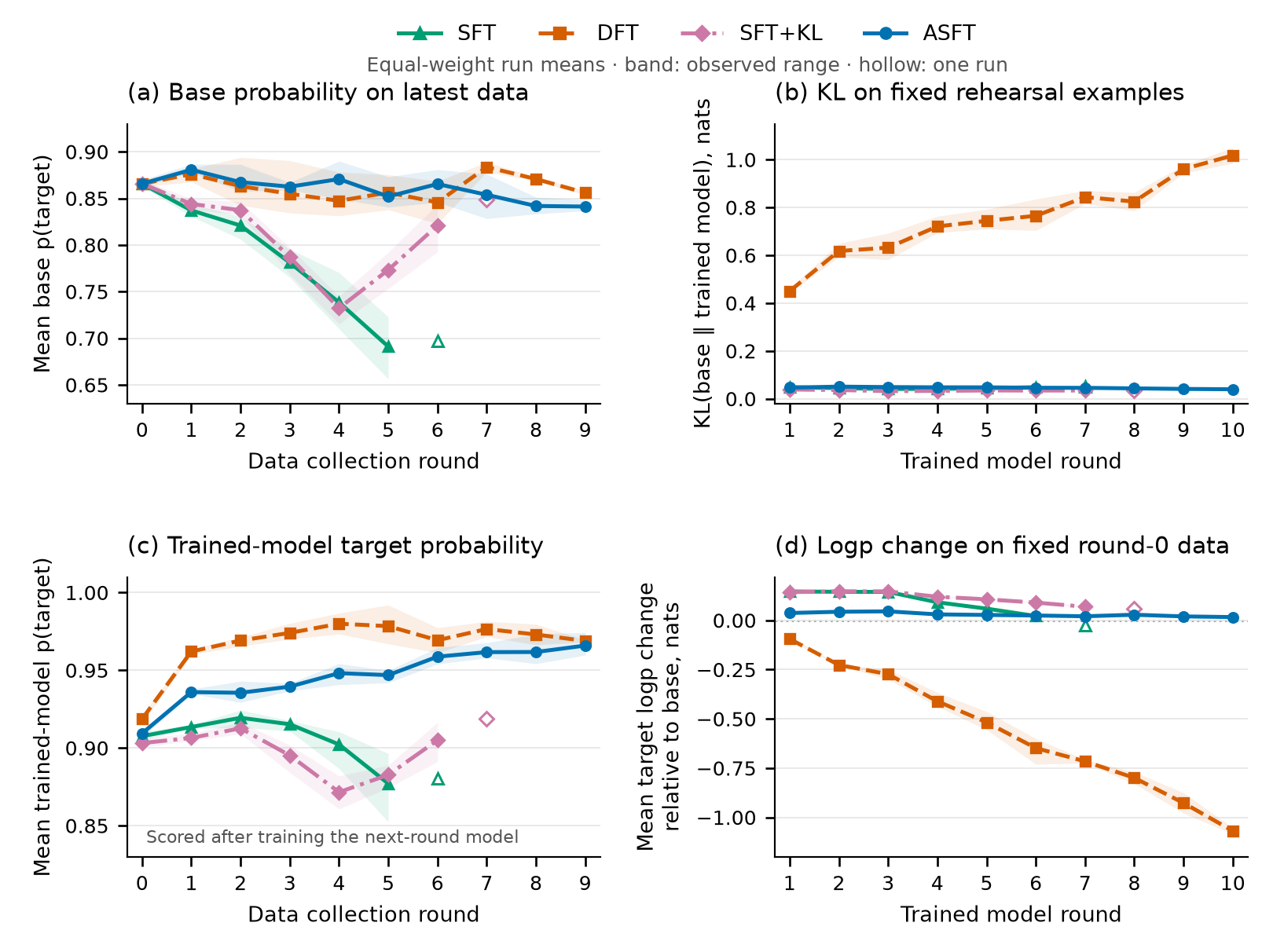}
\caption[Training-example probabilities and model comparisons during recursive training]{\textbf{Training-example probabilities and model comparisons during recursive training.} (a) Mean base-model probability on target tokens in each lineage's newest selected examples. (b) Mean reference-to-current KL(base \ensuremath{\Vert} trained model) over full next-token distributions on a fixed rehearsal probe. (c) Mean trained-model probability on the same newest targets as in (a), measured after fitting. Data collected at round r train the round-(r+1) model, so the collection-round axes in (a) and (c) are offset by one from the trained-model-round axis in (b). (d) Mean target-token log-probability change from the base on a fixed round-0 sample. This round-0 sample remains in the data buffer till round 3 and is absent from round 4 onward. Curves show equal-weight means across observed runs; bands show observed minimum--maximum ranges. Filled markers require at least two contributing runs, and disconnected hollow markers denote single-run measurements. Lines connect measured means. Scoring, coverage, and token-weight diagnostics are described in \hyperref[sec:A.7]{Appendix~A.7}.}
\label{fig:6}
\end{figure}

ASFT sustains successful recursion across all three runs (\hyperref[fig:4]{Fig.~4}). It also improves on DFT when both objectives fit the same recorded buffers (\hyperref[fig:5]{Fig.~5a}). These results motivate examining how probability weighting interacts with anchoring to the base. \hyperref[fig:6]{Figure~6} compares the fit to newly selected targets and tracks how predictions on fixed rehearsal and round-0 examples change relative to the base.

ASFT assigns high mean probability to newly selected target tokens after fitting (\hyperref[fig:6]{Fig.~6c}), while its rehearsal-probe KL stays small and its mean target log probability on the fixed round-0 sample stays close to the base model's (\hyperref[fig:6]{Fig.~6b} \& 6d). DFT assigns higher mean probability to its newest targets (\hyperref[fig:6]{Fig.~6c}), while its rehearsal-probe KL grows and its mean round-0 target log probability falls progressively below the base model's (\hyperref[fig:6]{Fig~6b} \& 6d). SFT and SFT+KL also stay close to the base on rehearsal contexts (\hyperref[fig:6]{Fig.~6b}). In both SFT and SFT + KL, the base assigns progressively lower probabilities to the newest targets through collection round 4 (\hyperref[fig:6]{Fig.~6a}).

Rehearsal-probe KL compares a fitted policy with the base on fixed contexts. The fitting sensitivity in \hyperref[prop:1]{Proposition~1} bounds how much the fitted policy changes when the training distribution changes.

Both DFT and ASFT scale the target-fitting gradient by the current target probability $p_y$, attenuating the signal for low-probability targets. ASFT also adds a forward-KL gradient toward the base. For the same KL coefficient $\lambda$, the ratio of the reference coefficient to the target-fitting coefficient is $\lambda/p_y$ in ASFT and $\lambda$ in SFT+KL (\hyperref[sec:A.3]{Appendix~A.3}). The measured training gradients follow a pattern consistent with the probability weighting. Across model rounds 1--6, mean pre-clipping parameter-gradient norms stay low and approximately flat for DFT and ASFT, while they rise for SFT and SFT+KL (\hyperref[fig:A6]{Fig.~A6}).

This interaction between probability weighting and anchoring suggests a mechanism for ASFT's pattern in \hyperref[fig:6]{Figure~6}. When the two effects are combined, they may moderate how strongly low-probability targets redirect the fit, while reinforcing behavior where the model already assigns greater probability.

These diagnostics follow ``pure lineages'' where each differs in both training objective and data buffer. When a DFT lineage is switched to ASFT, the first ASFT fit would use the existing DFT-produced buffer. \hyperref[sec:4.4]{\S{}4.4} follows task and collection success after the switch. \hyperref[sec:4.5]{\S{}4.5} compares the resulting buffers with those of matched ASFT-throughout lineages.

\subsection{Objective switches and buffer turnover}\label{sec:4.4}

In \hyperref[sec:4.2]{\S{}4.2}, we showed that using the same data but changing the training objective can lead to improved task success. Here, we follow objective switches across later rounds to see how task and collection success change as data generated after the switch replace the inherited data.

\begin{figure}[htbp]
\centering
\includegraphics[width=\linewidth,height=.74\textheight,keepaspectratio]{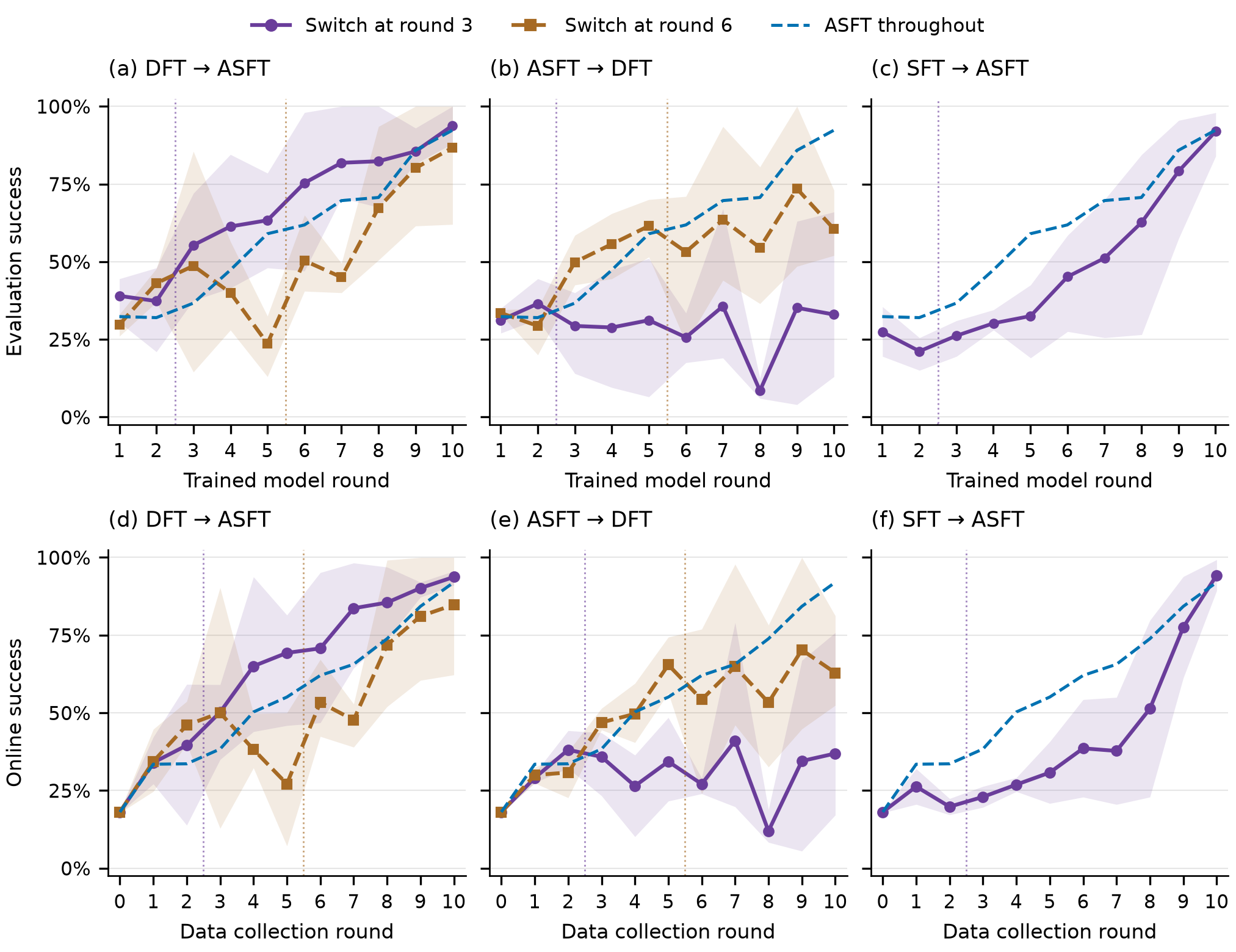}
\caption[Switches to ASFT finish with high success after buffer turnover]{\textbf{Switches to ASFT finish with high success after buffer turnover.} Panels (a--c) show success on the fixed 200-item evaluation panel. Panels (d--f) show online collection success, with round 0 generated by the base model. The columns show DFT\ensuremath{\rightarrow}ASFT (a,d), ASFT\ensuremath{\rightarrow}DFT (b,e), and SFT\ensuremath{\rightarrow}ASFT (c,f). Purple circles and brown squares identify switches at model rounds 3 and 6, respectively. Dotted vertical lines separate the rounds before and after each switch. The first fit whose retained collections all come from post-switch models is at round 6 for a round-3 switch and round 9 for a round-6 switch. Each curve shows equal weighted averages, while shaded bands show the observed minimum-maximum ranges. \hyperref[fig:A2]{Fig.~A2} shows the individual fixed-panel trajectories. The blue dashed curves show mean success for pure-ASFT lineages.}
\label{fig:7}
\end{figure}

At the switch round, the new objective fits a fresh adapter from the common base using the buffer inherited from the original lineage. We test DFT\ensuremath{\rightarrow}ASFT and ASFT\ensuremath{\rightarrow}DFT beginning at model round 3 or 6, and SFT\ensuremath{\rightarrow}ASFT beginning at model round 3 (\hyperref[fig:7]{Fig.~7}).

All three schedules that switch to ASFT reach high fixed-panel success by round 10. DFT\ensuremath{\rightarrow}ASFT reaches 93.8\% fixed-panel and 93.7\% online success when the switch begins at round 3, and 86.7\% and 84.8\% when it begins at round 6 (\hyperref[fig:7]{Fig.~7a},d). SFT\ensuremath{\rightarrow}ASFT improves unevenly from 45.2\% fixed-panel success at round 6 to 92.0\% at round 10 (\hyperref[fig:7]{Fig.~7c}). ASFT throughout reaches 92.3\% fixed-panel and 92.0\% online success. The reverse ASFT\ensuremath{\rightarrow}DFT switches finish with lower success on both measures. The round-3 switch reaches 33.0\% fixed-panel and 36.9\% online success, and the round-6 switch reaches 60.5\% and 62.9\% (\hyperref[fig:7]{Fig.~7b},e).

Consider the DFT\ensuremath{\rightarrow}ASFT switch at model round 6. The first ASFT fit uses the three collections generated by the DFT models at rounds 3--5. The round-6 ASFT model then generates the first post-switch collection, which enters the buffer for the round-7 fit. By model round 9, all three retained collections come from post-switch ASFT-fitted checkpoints at rounds 6--8. For a switch at model round 3, a similar state is reached at round 6. The high round-10 success rates for switches to ASFT therefore occur after all pre-switch collections have left the buffer.

Inherited collections can shape the data that replace them. A model fitted just after the switch may learn a cleanup procedure from the inherited data and use it on a successful attempt. If that attempt is selected, the new record can carry the procedure forward after the original collection leaves. \hyperref[sec:4.5]{\S{}4.5} probes whether paired buffers become more similar over complete turnovers under ASFT.

\subsection{Buffer separation and behavioral narrowing}\label{sec:4.5}

\hyperref[prop:1]{Proposition~1} predicts contraction over one complete buffer turnover when two lineages follow the same fixed-mixture population update and its contraction factor is below one. Write $\Delta_t$ for the distance between their buffers at model round $t$. Under these conditions, a shared ASFT update gives $\Delta_{t+3}\leq q_{\mathrm{ASFT}}\Delta_t$. We paired each DFT\ensuremath{\rightarrow}ASFT lineage with the ASFT-throughout lineage from the same starting dataset and measured $\Delta_t$ as the maximum Wasserstein distance between selected collections of the same age (\hyperref[sec:A.10]{Appendix~A.10}).

\begin{figure}[htbp]
\centering
\includegraphics[width=\linewidth,height=.74\textheight,keepaspectratio]{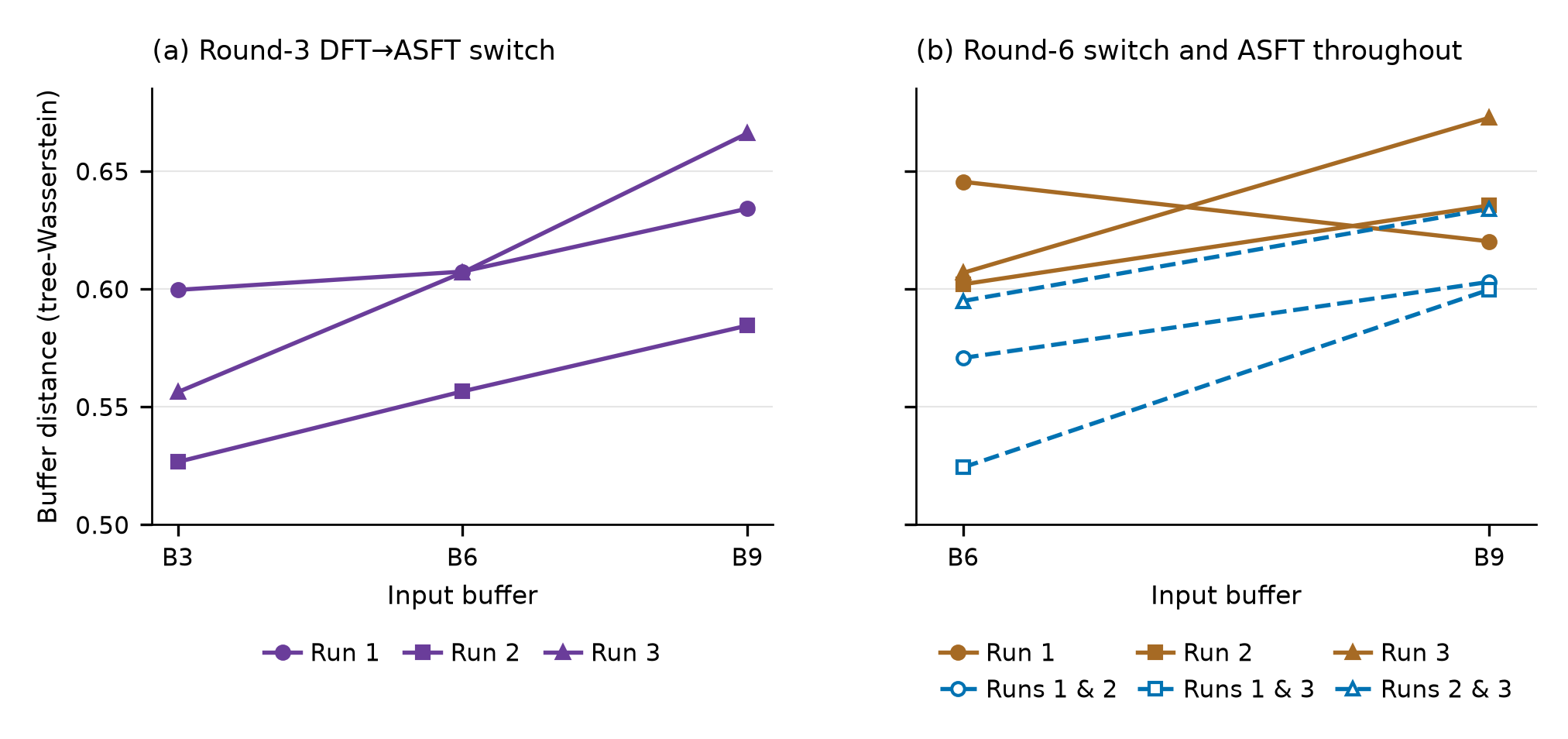}
\caption[Between-lineage buffer distance across complete turnovers under ASFT]{\textbf{Between-lineage buffer distance across complete turnovers under ASFT.} (a) Each round-3 DFT\ensuremath{\rightarrow}ASFT lineage and its ASFT-throughout reference from the same starting dataset, measured at the switch and after one and two turnovers. (b) Round-6 switch pairs and pairs of ASFT-throughout lineages over rounds 6--9. Each point is the maximum tree-Wasserstein distance across the three matching collection ages, with equal mass per valid target-token occurrence.}
\label{fig:8}
\end{figure}

For the round-3 switch, the mean buffer distance rose from 0.561 at the switch to 0.590 after one turnover and 0.628 after two (\hyperref[fig:8]{Fig.~8a}). All three pairs widened over both intervals under token weighting. Giving each episode equal mass or fixing the task-variant mix preserved the second increase in every pair. By round 6, every retained collection in both lineages had been generated under ASFT. Over the next turnover, the distance also increased in every pair when we pooled the three collections with rehearsal at each fit's realized token shares (mean 0.532 to 0.576, \hyperref[tab:A3]{Table~A3}). At round 9, the switched and ASFT-throughout lineages nevertheless achieved nearly identical mean fixed-panel success across the three runs (85.5\% and 85.8\%, respectively, \hyperref[fig:7]{Fig.~7a}). For the round-6 switch, two pairs widened and one narrowed. All three pairs of ASFT-throughout lineages widened over the same interval (\hyperref[fig:8]{Fig.~8b}).

At round 8, mean fixed-panel success was 67.3\% for the lineages that switched from DFT to ASFT at round 6 and 70.7\% for those trained with ASFT throughout. Passing required agents to clear at least 5\% of disposable files and satisfy the integrity and process checks. On items that both agents passed, mean clearance was 53.8\% of disposable fixture paths after the switch and 31.0\% under ASFT throughout, with each run weighted equally. Even among these shared successes, the switched agents removed substantially more of what could safely be cleared.

\begin{figure}[htbp]
\centering
\includegraphics[width=\linewidth,height=.74\textheight,keepaspectratio]{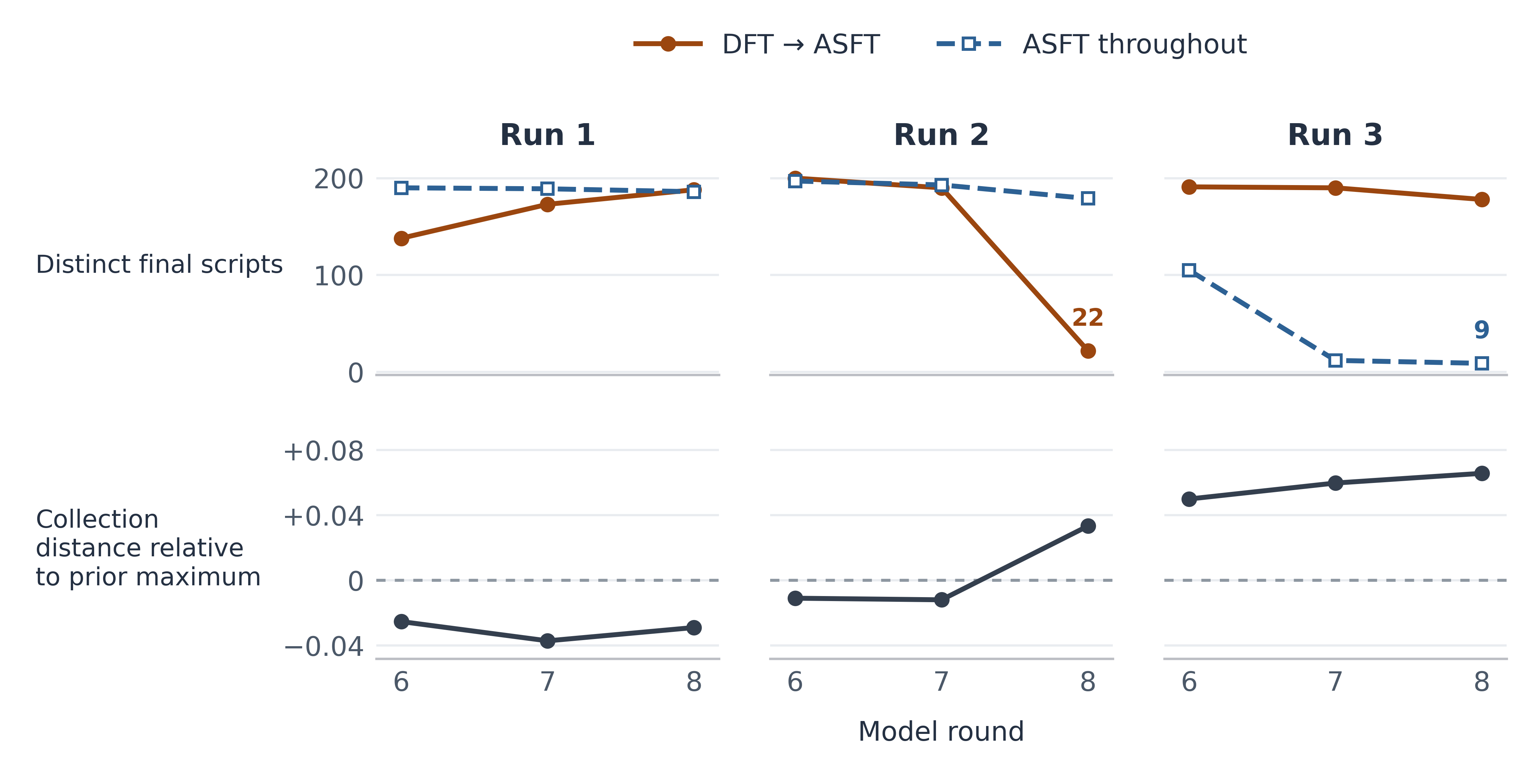}
\caption[Final-script diversity and separation between newly selected collections after the round-6 switch]{\textbf{Final-script diversity and separation between newly selected collections after the round-6 switch.} Each column pairs a switched lineage with the ASFT-throughout lineage from the same starting dataset. Top, number of distinct final script bodies across all 200 fixed-panel items. Bottom, token-weighted tree-Wasserstein distance between the two collections generated at each round, minus the largest age-matched distance in that pair's round-6 input buffers. Above zero, a replacement collection pair exceeds that earlier maximum. Only the round-8 pair exceeded it in Run 2, whereas all three replacement pairs did in Run 3.}
\label{fig:9}
\end{figure}

In Run 2, the switched lineage's success rose from 80 to 187 of 200 items between rounds 7 and 8 as its repertoire fell from 190 distinct final script bodies to 22. Among the 73 items it passed at both rounds, the number with removal reports during exploration rose from 60 to 73, while the number with such reports during cleanup fell from 11 to zero. Its ASFT-throughout counterpart still produced 179 distinct final bodies at round 8. In Run 3, the ASFT-throughout lineage narrowed to nine distinct bodies while the switched lineage produced 178. The paired buffer distance widened in both runs (\hyperref[fig:9]{Fig.~9}).

For the round-3 switch, \hyperref[prop:1]{Proposition~1} would require $\Delta_9<\Delta_6$ if the lineages followed a common contractive population update. We observed $\Delta_9>\Delta_6$ in every pair over a turnover that began after the inherited DFT collections had left the buffer. The observed distances leave the population contraction factor undetermined. Each realized fit uses finite collections whose token counts set the age and rehearsal weights, and the controls reweight the archived data only after fitting. Extending the argument would require bounding how changes in both token distributions and token counts propagate through fitting and collection. On this task, high success under a shared objective was compatible with increasing separation in the data passed to the next fit.

\subsection{Future Work: successful recursion and dynamical attraction}\label{sec:4.6}

Continued collection is an operational requirement of the recursive process. SFT and SFT+KL stop when their collections yield too few selected rows to continue, even though their recorded buffers can support more successful agents under alternative fits.

Collection may also contribute to sensitivity. Even with a fixed success rule, the selected-attempt distribution conditions a policy-dependent attempt distribution on success. When success is rare, changes in the attempt distribution can be magnified by this conditioning. Conversion into rows and target tokens further weights successful attempts by their recorded contributions. This is a possible amplification mechanism; the current stopping outcomes establish inadequate collection yield, rather than measuring that amplification.

A focused follow-up is to perturb a recorded buffer and continue the perturbed and unperturbed lineages under the same objective. Policy and selected-data differences over complete window turnovers would assess whether the disturbance is damped, while changes within each trajectory would assess whether the process is settling. Task success and selected-row yield would track performance and collection productivity, while program identity would show which executable outputs recur.

\subsection{Limitations}\label{sec:4.7}

The recursive experiments cover one model, one task family, three starting datasets with a common training seed, and a finite horizon. FictionalQA adds single-round document adaptation across three training seeds, with its evaluation and scoring scope described in \hyperref[sec:A.9]{Appendix~A.9}. The crossed matrix compares realized fits, including historical diagonals, and the switch schedules follow separate pre-switch histories. While we would have liked to explore optimal objective switching time or recovery, these would require more extensive and varied tests.

\FloatBarrier

\section{Conclusion}\label{sec:5}

In recursive self-training, the choice of training objective affects both the agent fitted in the current round and the examples available to later agents. On the file-cleanup task, DFT led in online success early in training, while ASFT sustained collection through all ten rounds across all three starting datasets. The round-6 refits show why a checkpoint score gives only a partial account of its lineage. The original DFT fits achieved 22.2\% mean success on the fixed evaluation panel. Fitting fresh adapters to the same data with ASFT raised mean success to 66.8\%.

The switch experiments reveal a complementary distinction. By round 9, lineages that switched from DFT to ASFT at round 3 achieved 85.5\% mean success, compared with 85.8\% for those trained with ASFT throughout. Yet the distance between the retained data of every matched pair increased from rounds 6 to 9, an interval that began after all DFT-generated collections had left the buffer. Similar task performance was therefore compatible with growing differences in what the agents passed forward. We therefore suggest evaluating a recursive training objective both by how well its agents perform and by whether the examples they collect are sufficient to sustain training and useful to their successors.

\clearpage
\appendix
\setcounter{figure}{0}
\setcounter{table}{0}
\renewcommand{\thefigure}{A\arabic{figure}}
\renewcommand{\thetable}{A\arabic{table}}
\renewcommand{\theHfigure}{app.\arabic{figure}}
\renewcommand{\theHtable}{app.\arabic{table}}
\section{Appendix}\label{sec:A}

\subsection{Proof of Proposition 1}\label{sec:A.1}

Use the same coupling for the shared rehearsal component and couple corresponding age components. Convexity of $W_1$ under common mixture weights gives

\[
\begin{aligned}
W_1(P(\mathbf{B}),P(\mathbf{B}'))
&\leq(1-\rho)\sum_{j=1}^{K}a_jW_1(Q_j,Q_j')\\
&\leq(1-\rho)d_B(\mathbf{B},\mathbf{B}').
\end{aligned}
\]

Thus the new-collection map $H_u=C\circ F_u\circ P$ obeys

\[
W_1(H_u(\mathbf{B}),H_u(\mathbf{B}'))
\leq q_u d_B(\mathbf{B},\mathbf{B}').
\]

The remaining coordinates of $\Phi_u$ shift existing entries. Hence $\Phi_u$ is nonexpansive for $q_u<1$. If the initial buffer distance is $E$, every collection generated during the next $K$ steps differs by at most $q_uE$: intermediate buffer distances never exceed $E$. After $K$ shifts, every original entry has left the buffer, proving the claimed $K$-step contraction.

Banach's fixed-point theorem on the complete invariant domain gives a unique fixed point of $\Phi_u^K$. Because $\Phi_u$ commutes with its $K$th iterate, applying $\Phi_u$ to that point gives another fixed point of $\Phi_u^K$; uniqueness makes it a fixed point of $\Phi_u$ itself. The shift equations force all entries to be identical. Repeated block contraction, with nonexpansiveness for the remaining steps, gives the stated convergence rate. Fitted-policy convergence follows from the assumed Lipschitz bound for $F_u$.

Completeness refers, for example, to a closed invariant subset of the finite-first-moment probability measures on a complete separable ground space, with the product maximum metric. Invariance is substantive: collection and fitting must remain defined on the domain. A local version requires such a closed invariant neighborhood and an initial buffer within it.

\subsection{Bounded collection and fitting errors}\label{sec:A.2}

Assume the realized buffers remain in the admissible domain. Suppose the realized new collection has a uniformly bounded discrepancy from the ideal update on its own current buffer:

\[
W_1(\widehat Q_t,H_u(\widehat{\mathbf{B}}_t))\leq\eta.
\]

For example, a policy-fitting error at most $\varepsilon_{\mathrm{fit}}$ and a collection-law error at most $\varepsilon_{\mathrm{collect}}$ imply $\eta\leq L_{\mathrm{collect}}\varepsilon_{\mathrm{fit}}+\varepsilon_{\mathrm{collect}}$. Any bounded discrepancy from using an idealized rather than realized mixture can also be included in $\eta$ if a common comparison map is specified. The error bounds are assumed inputs to the analysis.

Let $e_t=W_1(\widehat Q_t,Q_u^*)$ and $E_t=d_B(\widehat{\mathbf{B}}_t,\mathbf{B}_u^*)$. The triangle inequality gives

\[
e_t\leq q_u\max_{1\leq j\leq K}e_{t-j}+\eta.
\]

Set $r=\eta/(1-q_u)$ and $x_+=\max\{x,0\}$. Subtracting $r$ gives

\[
(e_t-r)_+\leq q_u\max_{1\leq j\leq K}(e_{t-j}-r)_+.
\]

The same window-turnover argument therefore yields

\[
(E_{t+K}-r)_+\leq q_u(E_t-r)_+,
\qquad
\limsup_{t\to\infty}E_t\leq\frac{\eta}{1-q_u}.
\]

Under persistent bounded error, the buffer's distance from the ideal fixed point satisfies $\limsup_{t\to\infty}E_t\leq\eta/(1-q_u)$. For stochastic errors, this is a pathwise conclusion on an event where the uniform error bound holds. A finite-horizon high-probability claim would additionally require a joint probability bound on that event. Training rows from the same successful attempt are also dependent.

\subsection{Local gradients and curvature}\label{sec:A.3}

\subsubsection*{Local interaction of weighting and anchoring}

At one recorded prefix, let $p=\operatorname{softmax}(z)$ be the predicted vocabulary distribution, $p_0$ the fixed base distribution, $y$ the target token, and $e_y$ its one-hot vector. With detached probability weighting and the forward divergence $D_{\mathrm{KL}}(p_0\Vert p)$, the logit gradients are

\[
\begin{aligned}
\nabla_z\ell_{\mathrm{SFT}} &= p-e_y,\\
\nabla_z\ell_{\mathrm{DFT}} &= p_y(p-e_y),\\
\nabla_z\ell_{\mathrm{SFT+KL}} &= (p-e_y)+\lambda(p-p_0),\\
\nabla_z\ell_{\mathrm{ASFT}} &= p_y(p-e_y)+\lambda(p-p_0).
\end{aligned}
\]

At fixed prediction, probability weighting scales the norm of the target-fitting gradient by $p_y$. With the same $\lambda$, the ratio of the KL coefficient to the target-fitting coefficient is $\lambda/p_y$ in ASFT and $\lambda$ in SFT+KL. The relative coefficient on the KL term in ASFT therefore increases as the target probability decreases.

\subsubsection*{Curvature in probability coordinates}

At a fixed prefix, let $p$ be a free probability vector over the finite vocabulary, $p_0$ its base reference distribution, and $r$ the empirical target distribution at that prefix. The ASFT potential is

\[
\widetilde\ell(p;r,p_0)
=-\sum_v r_vp_v+\lambda\sum_v p_{0,v}\log\frac{p_{0,v}}{p_v}.
\]

The target-fitting term is linear in $p$. In the interior of the simplex, the Hessian comes entirely from the forward-KL term:

\[
\nabla_p^2\widetilde\ell
=\lambda\operatorname{diag}\!\left(\frac{p_{0,v}}{p_v^2}\right).
\]

With $\lambda>0$ and a full-support reference, this is positive definite on every nonzero tangent direction of the simplex. The KL term therefore supplies the curvature in probability coordinates. This probability-coordinate curvature is shared by ASFT and SFT+KL.

\subsection{Per-lineage recursive outcomes}\label{sec:A.4}

\begin{figure}[htbp]
\centering
\includegraphics[width=\linewidth,height=.74\textheight,keepaspectratio]{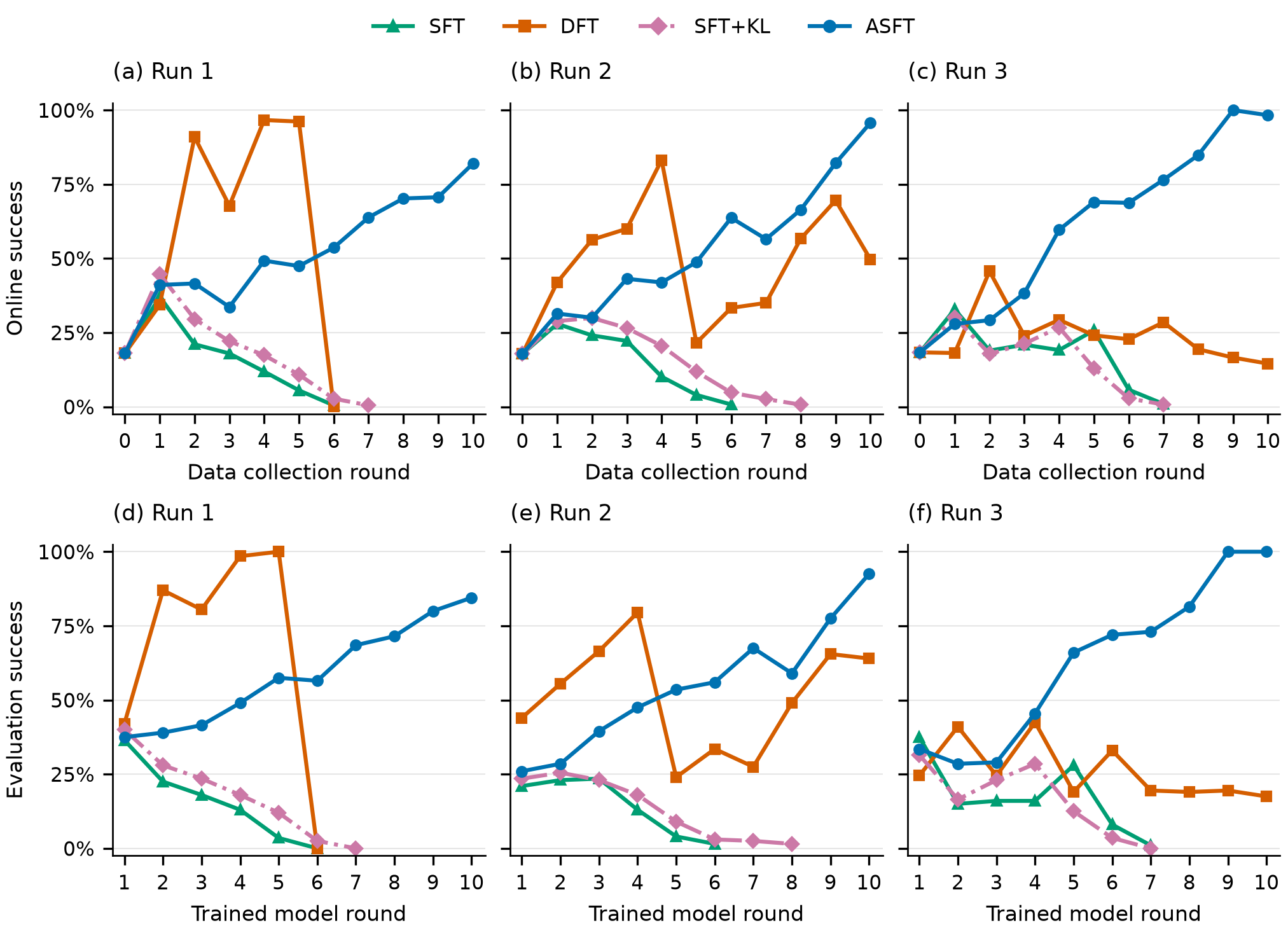}
\caption[ASFT continues consistently across starting datasets]{\textbf{ASFT continues consistently across starting datasets.} Columns identify the three starting-data lineages. (a--c) Online collection success for the four training objectives. (d--f) Fixed-panel evaluation success for the available checkpoints. Panel (d) uses the dedicated reevaluation at round 6 and the August sweep at other rounds. Data collected at round g enter training for model round g + 1. These individual runs underlie both panels of \hyperref[fig:4]{Fig.~4}.}
\label{fig:A1}
\end{figure}

\hyperref[fig:A1]{Figure~A1} separates the three starting-data runs underlying \hyperref[fig:4]{Fig.~4}. Each run begins with a distinct collection generated by the base policy; within a run, the four objectives share that starting collection. The runs describe variation across starting-data histories at a fixed training seed.

Fixed-panel success is measured on the common 200-item evaluation panel. The collection target counts selected training rows, and one successful attempt can provide several rows; it is therefore distinct from the number of successful attempts. Data collected at round $g$ first enter the fit for model round $g+1$. Each fit starts a fresh adapter from the common pretrained base.

ASFT continues through round 10 in every run, whereas SFT and SFT+KL stop between rounds 6 and 8 after collecting too few selected rows to continue. DFT has heterogeneous outcomes: Run 1 stops at round 6, while Runs 2 and 3 continue through round 10 with substantially different evaluation success. These individual trajectories show both the consistent ASFT continuation and the variation hidden by an aggregate DFT curve.

An evaluated checkpoint and its ability to collect enough data for another fit are separate observations. The late round DFT aggregation in \hyperref[fig:4]{Fig.~4} hence describes only its two continuing runs. Throughout the paper, observed run ranges are descriptive minimum--maximum ranges, rather than confidence intervals.

\subsection{Per-lineage objective-switch outcomes}\label{sec:A.5}

\begin{figure}[H]
\centering
\includegraphics[width=.85\linewidth,height=.74\textheight,keepaspectratio]{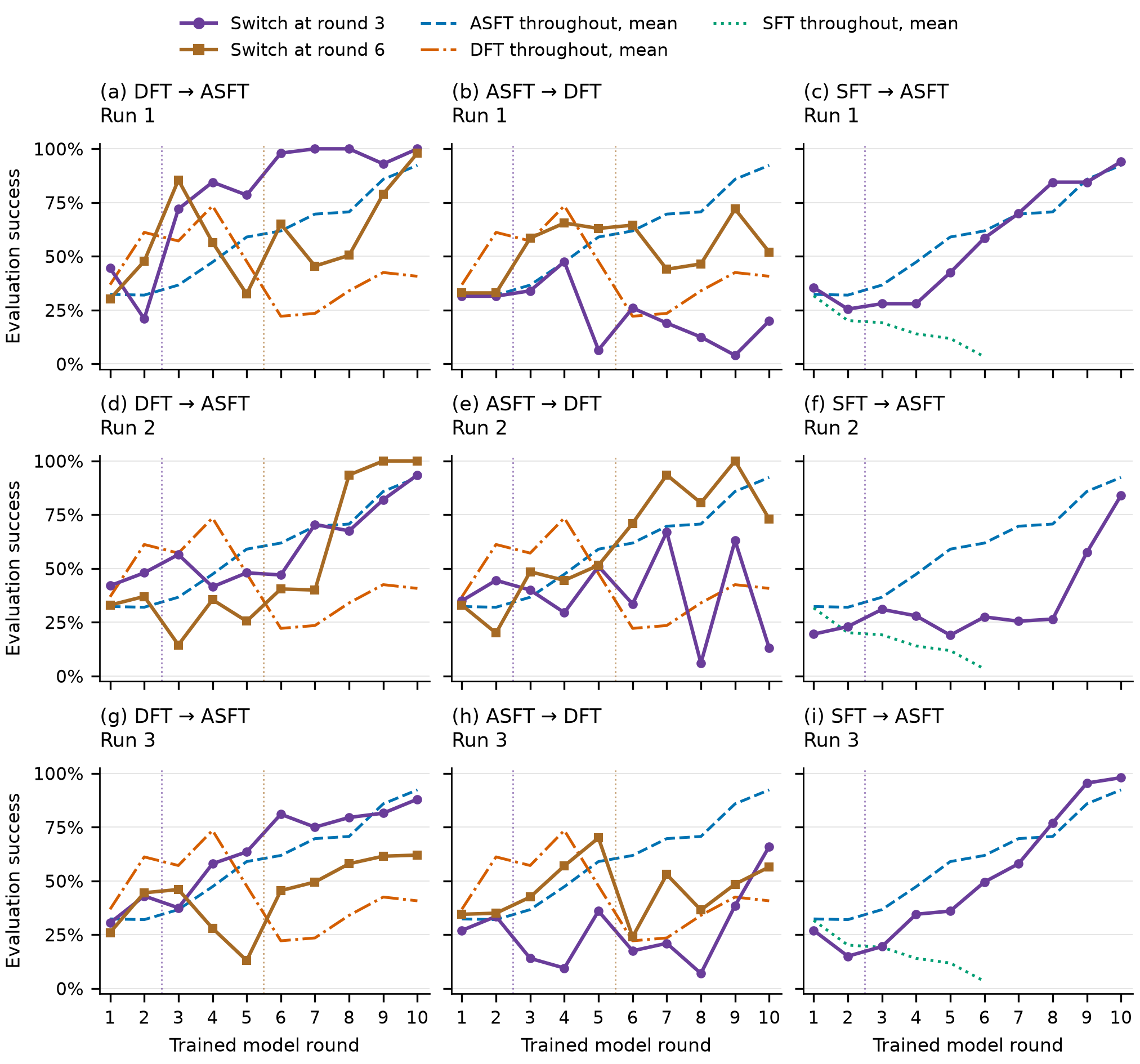}
\caption[Switches to ASFT support high later success across starting-data histories]{\textbf{Switches to ASFT support high later success across starting-data histories.} Rows identify Runs 1--3; columns show DFT\ensuremath{\rightarrow}ASFT, ASFT\ensuremath{\rightarrow}DFT, and SFT\ensuremath{\rightarrow}ASFT. Solid curves show individual switched lineages, with purple circles for switches beginning at model round 3 and brown squares for round 6. Dotted lines separate the final pre-switch fit from the first fit under the new objective. Dashed, dash-dotted, and dotted reference curves show the means across available runs for constant ASFT, DFT, and SFT, respectively, repeated across rows. Every checkpoint is evaluated on the common 200-item panel. Each fit trains a fresh adapter from the common base while carrying retained data forward.}
\label{fig:A2}
\end{figure}

\hyperref[fig:A2]{Figure~A2} separates the objective-switch evaluation curves in \hyperref[fig:7]{Fig.~7} into their individual starting-data histories. We follow DFT\ensuremath{\rightarrow}ASFT and ASFT\ensuremath{\rightarrow}DFT switches beginning at model round 3 or 6, and SFT\ensuremath{\rightarrow}ASFT beginning at round 3. Each schedule follows its recorded pre-switch history. All fifteen switched lineages have evaluation checkpoints through round 10.

Switches to ASFT reach high later success after both SFT and DFT beginnings, with different paths across runs. The SFT\ensuremath{\rightarrow}ASFT trajectories show delayed improvement, especially in Run 2, while the reverse ASFT\ensuremath{\rightarrow}DFT switches produce more variable later outcomes. The three-collection window makes rounds 6 and 9 the first fits using entirely post-switch collections for switches beginning at rounds 3 and 6, respectively. The round-10 outcomes therefore follow complete handover of the directly retained buffer.

\subsection{Exact-source repetition in fixed-panel evaluations}\label{sec:A.6}

We analyzed archived generated responses from all 247 indexed checkpoint evaluations across 27 lineages: 97 constant-objective and 150 objective-switch evaluations, totaling 49,400 episodes. Each evaluation uses the same 200 item identities and inference seeds, with 40 items from each of five filesystem variations. Runs vary their starting selected datasets while sharing a training seed. Here M denotes the trained model round; a switch beginning at M6 first uses the new objective when fitting M6.

For each exploration or cleanup response, we select the first fenced block with an accepted Python, shell, or empty language tag and a nonempty body. We trim outer whitespace, preserving internal formatting, comments, identifiers, literals, and statement order. The final-script measurement uses the last extracted body across both phases, compared by exact source equality. Both phases matter: all 400 Run 2 DFT\ensuremath{\rightarrow}ASFT-from-M6 episodes at M9--M10 perform deletion during exploration, which also influences their final extracted code bodies.

Complete code sequences retain every extracted body and its phase in recorded response order, including repeated calls. Response transcripts retain all generated model responses in order, including prose. These measurements differ: Run 3 ASFT uses the same complete code sequence across all items at both M9 and M10, with two response transcripts at each checkpoint. Run 2's persistent switch case has two final bodies but three sequences at M9 because one episode repeats a deletion call.

\begin{figure}[H]
\centering
\includegraphics[width=\linewidth,height=.65\textheight,keepaspectratio]{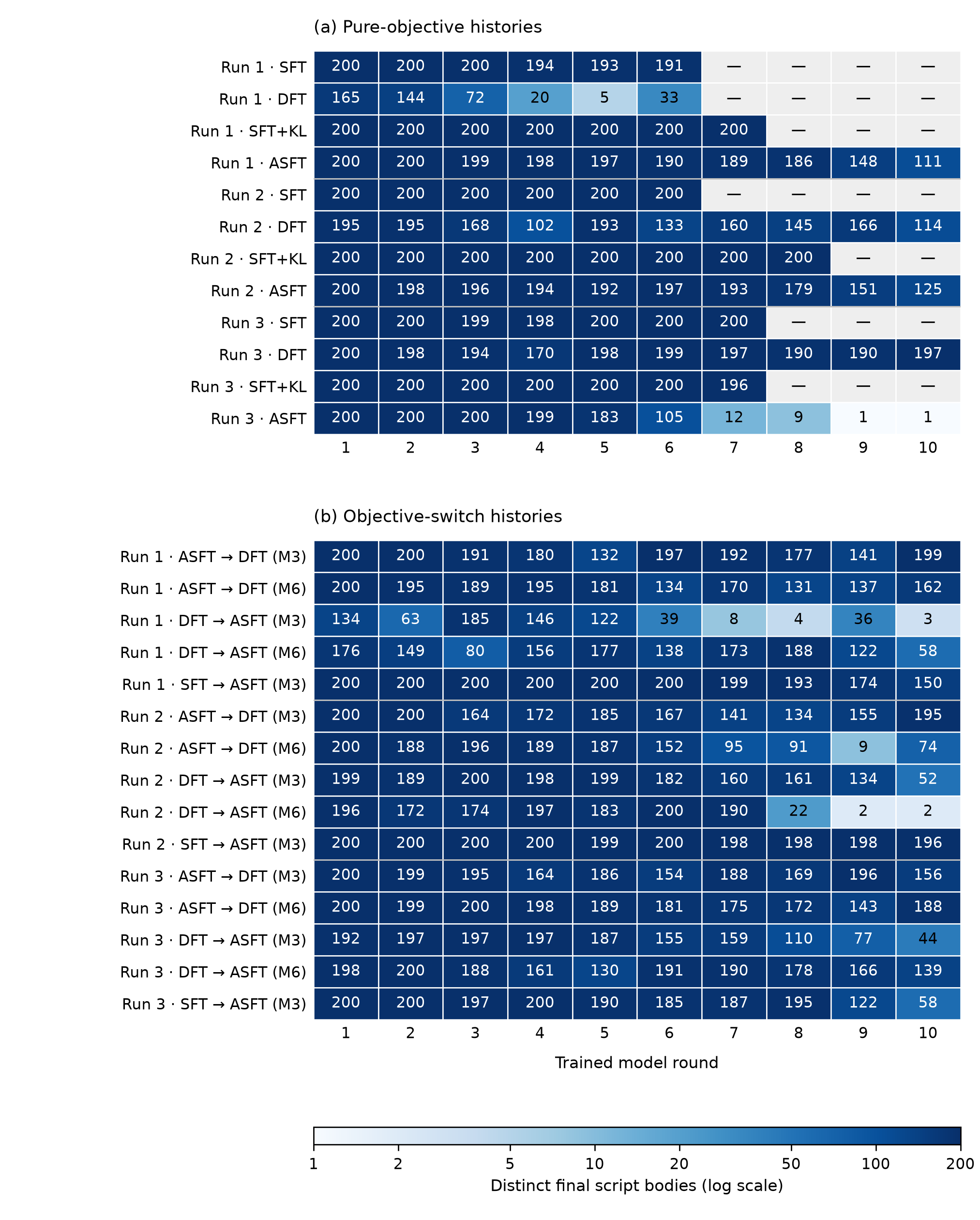}
\caption[Final-script counts for all 12 constant-objective histories]{\textbf{Final-script counts for all 12 constant-objective histories.} (a) and all 15 objective-switch histories (b). Cell labels give exact counts; the shared logarithmic color scale assigns darker shades to larger counts. A dash denotes no evaluation.}
\label{fig:A3}
\end{figure}

\hyperref[fig:A3]{Figure~A3} provides final-script counts across all indexed histories.

Four Run 3 SFT+KL M7 episodes have no extracted code body. Script and nonempty-sequence counts use the 196 program-bearing items; transcript and success measurements retain all 200 items.

\subsubsection*{Cleanup behavior around the round-6 switch}

In the Run 2 switched lineage, only one of the 200 evaluation items retained the same final code body from model round 7 to 8. Among the 73 items that succeeded at both checkpoints, the mean fraction of disposable fixture paths cleared rose from 31.5\% to 35.5\%. Phase-specific removal counts come from program reports in the execution traces. Clearance measures the fraction of disposable fixture paths absent at final measurement.

In Run 3, the ASFT-throughout lineage went from 105 distinct final bodies at round 6 to nine at round 8. Its switched counterpart produced 178 at round 8. The Run 1 switched lineage produced 138, 173, and 188 distinct final bodies at rounds 6--8, and its ASFT-throughout reference produced 190, 189, and 186.

\subsubsection*{Exact script recurrence across later fits}

\begin{figure}[H]
\centering
\includegraphics[width=\linewidth,height=.74\textheight,keepaspectratio]{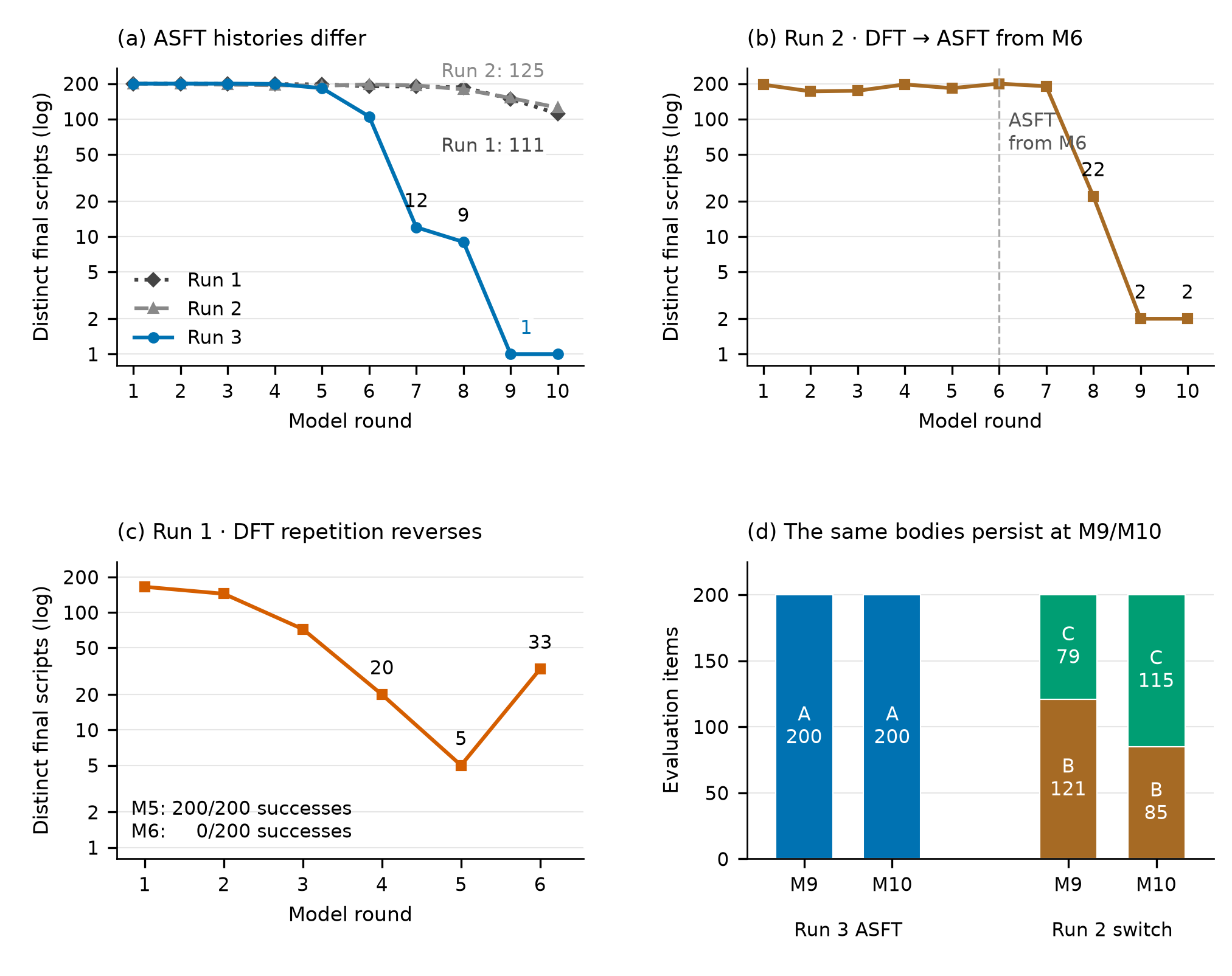}
\caption[Exact script repetition can persist or reverse across rounds]{\textbf{Exact script repetition can persist or reverse across rounds.} Distinct final executable script bodies on the fixed 200-item evaluation panel for (a) the three pure-ASFT lineages, (b) Run 2 DFT\ensuremath{\rightarrow}ASFT with ASFT beginning at round 6, and (c) Run 1 DFT. Count axes in (a--c) are logarithmic. (d) Body identities and frequencies at rounds 9 and 10 for the two persistent cases. Colors track body identity across rounds; all four evaluations in (d) achieve 200/200 successes.}
\label{fig:A4}
\end{figure}

At rounds 9 and 10, Run 3 ASFT used the same complete sequence of extracted code bodies across all 200 items. The Run 2 DFT\ensuremath{\rightarrow}ASFT lineage that switched at round 6 used the same two final bodies at both checkpoints, with frequencies changing from (121, 79) to (85, 115) (\hyperref[fig:A4]{Fig.~A4}). Both lineages succeeded on all 200 items at both checkpoints.

The other two ASFT-throughout lineages produced 111 and 125 distinct final bodies at round 10. Run 1 DFT went from five bodies and 200/200 successes at round 5 to 33 entirely different bodies and 0/200 successes at round 6.

\FloatBarrier

\subsection{Token and training-gradient diagnostics}\label{sec:A.7}

\subsubsection*{Token scoring and probability-weight dispersion}

The diagnostics in \hyperref[fig:6]{Fig.~6} distinguish changes in newly selected examples from changes in a fitted model's predictions. For the newest collection $A_g$, panels (a) and (c) score the same recorded target tokens under the base policy and the subsequently fitted policy $\pi_{g+1}$, respectively. Prefixes are held fixed during scoring. Each selected-data sample contains 256 examples, scored on completion tokens with a 16,384-token limit. Scoring uses temperature 1 and a bf16 proxy for the quantized reference.

\hyperref[fig:6]{Figure~6b} averages the forward divergence $D_{\mathrm{KL}}(\pi_0(\cdot\mid h)\Vert\pi_t(\cdot\mid h))$ over a fixed 500-example rehearsal probe, using up to 256 tokens per example and 56,105 scored positions. This compares full next-token distributions at recorded contexts. \hyperref[fig:6]{Figure~6d} instead averages paired target-token log-probability differences between the fitted policy and base on a fixed round-0 sample. It is a mean of log-probability differences, rather than the logarithm of a ratio of mean probabilities. That sample remains in the round-3 fitting window and leaves the window from round 4 onward.

The \hyperref[fig:6]{Fig.~6} measurements cover all 97 available checkpoints across the twelve constant-objective lineages. At each round, means weight contributing runs equally and bands show their observed minimum--maximum range. Filled markers require at least two runs; hollow, disconnected markers denote singleton endpoints. Changes in contributing runs reflect the observed lineage horizons, and no outcomes are imputed after stopping.

\begin{figure}[htbp]
\centering
\includegraphics[width=\linewidth,height=.74\textheight,keepaspectratio]{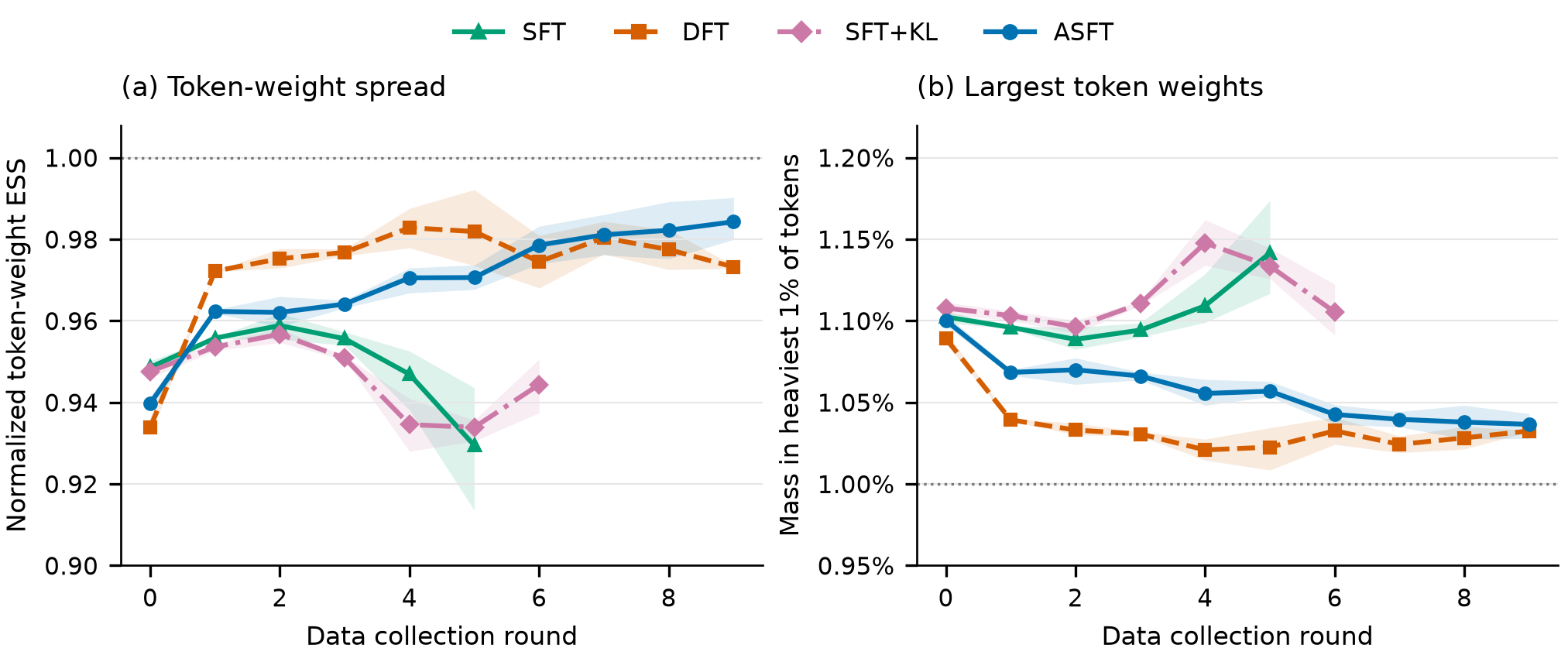}
\caption[Token-probability weights remain broadly distributed in the scored samples]{\textbf{Token-probability weights remain broadly distributed in the scored samples.} (a) Normalized token-weight ESS and (b) the share of total weight in the heaviest 1\% of tokens, measured on available newest selected-example samples also used for \hyperref[fig:6]{Fig.~6}. Weights are the fitted model's target-token probabilities after fitting those examples. Shading shows the observed minimum--maximum range. Gray lines indicate uniform weights. For SFT and SFT+KL, these are diagnostic probability weights rather than the unit weights used by their training objectives. The panels measure weight dispersion, rather than gradient concentration or update size.}
\label{fig:A5}
\end{figure}

\hyperref[fig:A5]{Figure~A5} examines the dispersion of target-probability weights on the scored newest selected examples. For valid target positions $i=1,\ldots,N$, define $w_i=\pi_{g+1}(y_i\mid h_i)$. The normalized token-weight effective sample size is

\[
\operatorname{ESS}_{\mathrm{norm}}
=\frac{\left(\sum_{i=1}^{N}w_i\right)^2}
{N\sum_{i=1}^{N}w_i^2}.
\]

The second statistic is the share of total weight assigned to the heaviest 1\% of scored tokens. Uniform weights give normalized ESS of one and a top-1\% mass of approximately 1\%. \hyperref[fig:A5]{Figure~A5} summarizes these statistics for the available scored samples.

Across the plotted samples, the probability weights remain broadly distributed. This observation concerns the relative dispersion of weights, and is compatible with changes in their absolute values or in support for particular historical targets. For SFT and SFT+KL, the plotted probabilities apply the DFT weighting rule only as a diagnostic (those objectives use unit target-token weights).

\subsubsection*{Training gradient norms}

\begin{figure}[htbp]
\centering
\includegraphics[width=\linewidth,height=.74\textheight,keepaspectratio]{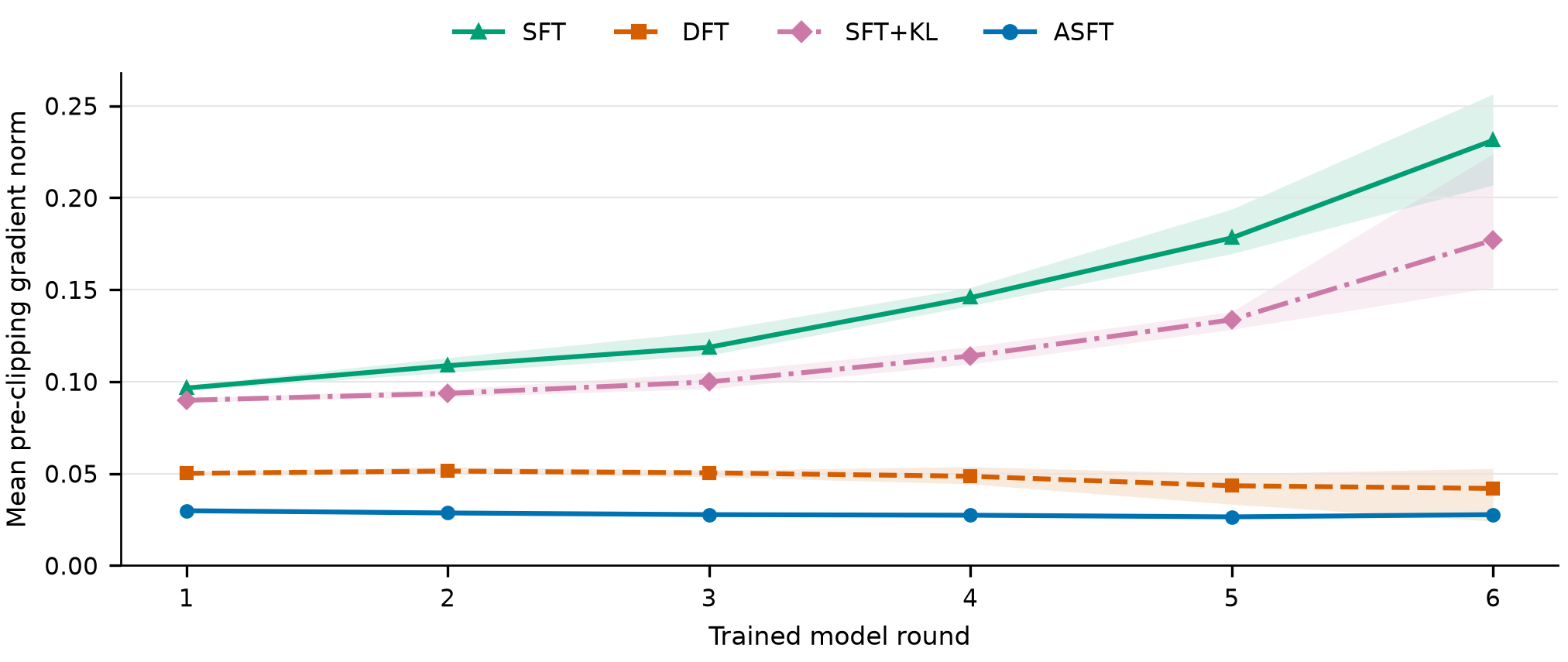}
\caption[Training gradient scale changes differently across objectives]{\textbf{Training gradient scale changes differently across objectives.} Mean pre-clipping gradient norm during adapter fitting at model rounds 1--6, averaged equally across three starting-data lineages. Shading shows their observed minimum--maximum range. Each round trains a fresh adapter from the same base. Norms combine selected data and rehearsal examples and summarize the realized training trajectories.}
\label{fig:A6}
\end{figure}

\hyperref[fig:A6]{Figure~A6} summarizes mean pre-clipping gradient norms recorded during adapter fitting at model rounds 1--6, the rounds available for every objective in all three starting-data lineages as outlined in \hyperref[sec:4]{\S{}4}. The measurements combine selected data and rehearsal examples.

ASFT has low, approximately constant mean gradient norms across these rounds, while SFT and SFT+KL show increasing norms. DFT also has low, approximately constant norms despite its poorer later evaluation performance. Low gradient magnitude alone therefore does not explain ASFT's sustained success. Beyond the shared initial buffer, the objectives follow different data histories; these curves describe the gradients encountered along their realized fitting trajectories rather than isolating an objective effect on a common buffer.

The measurements summarize combined training-gradient magnitudes.

\FloatBarrier

\subsection{Agreement during round-6 buffer fitting}\label{sec:A.8}

\begin{figure}[htbp]
\centering
\includegraphics[width=\linewidth,height=.74\textheight,keepaspectratio]{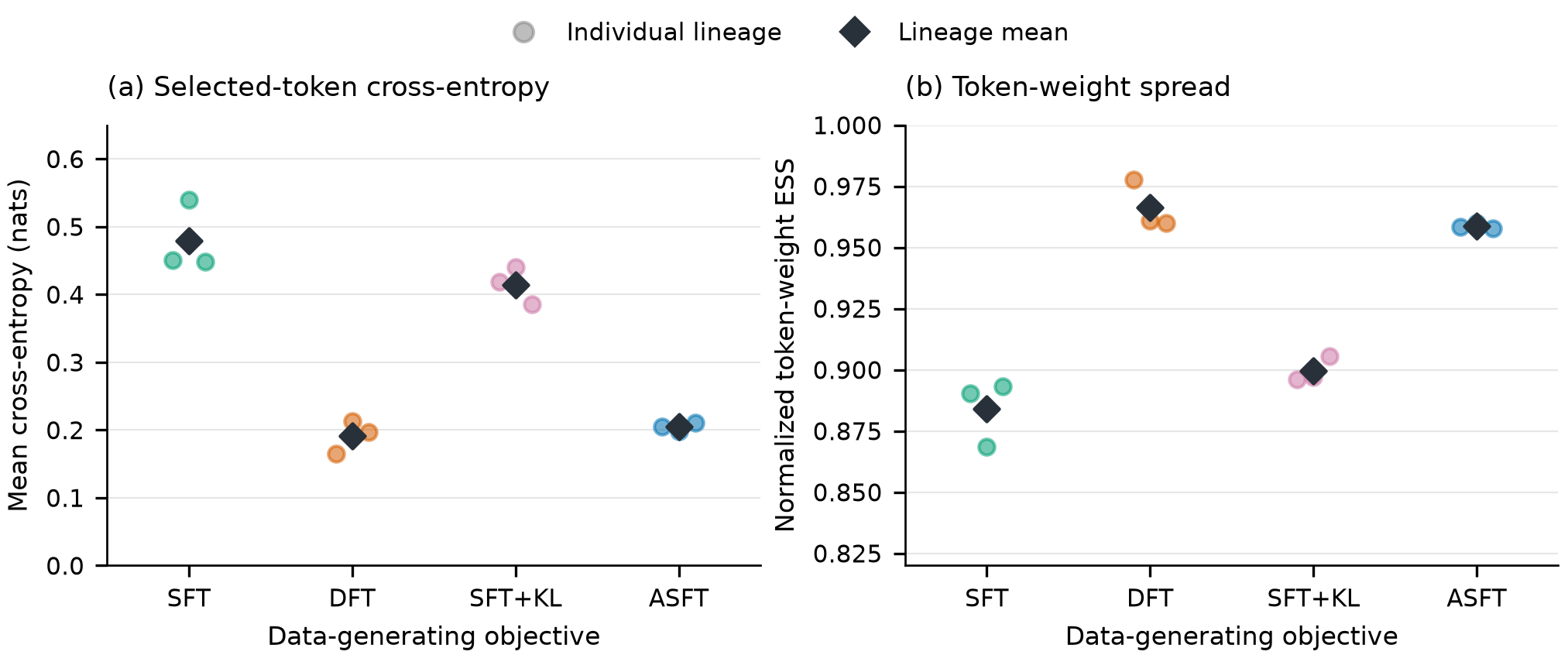}
\caption[Training-time agreement on round-6 training buffers]{\textbf{Training-time agreement on round-6 training buffers.} (a) Mean unweighted selected-token cross-entropy and (b) normalized token-weight ESS for ASFT fits of buffers generated by each source objective. Both statistics pool token moments within each fit before averaging across the three runs; rehearsal tokens are excluded.}
\label{fig:A7}
\end{figure}

\hyperref[fig:A7]{Figure~A7} uses ASFT as the training objective for each source objective's recorded round-6 training buffer. The horizontal axis identifies the objective used in the data-generating lineage. The measurements complement \hyperref[fig:5]{Fig.~5}'s evaluation outcomes by describing agreement with selected targets during fitting.

Let $\mathcal{I}$ index the valid selected-token occurrences scored throughout a fit, including repeated presentations as training proceeds, and let $p_i$ be the model's target probability when occurrence $i$ is scored. With $N_{\mathrm{train}}=|\mathcal{I}|$, the reported unweighted cross-entropy is

\[
\operatorname{CE}_{S}
=-\frac{1}{N_{\mathrm{train}}}\sum_{i\in\mathcal{I}}\log p_i.
\]

The training-time normalized ESS uses pooled weight moments:

\[
\operatorname{ESS}_{S,\mathrm{norm}}
=\frac{\left(\sum_{i\in\mathcal{I}}p_i\right)^2}
{N_{\mathrm{train}}\sum_{i\in\mathcal{I}}p_i^2}.
\]

Thus, token sums and counts are pooled within each fit before the three run-level statistics are averaged. Rehearsal tokens are excluded from both measurements. Both statistics aggregate observations collected throughout training. Normalized ESS is calculated from pooled token-weight moments rather than by averaging minibatch ESS values.

ASFT fits show higher unweighted selected-token cross-entropy and lower normalized ESS on SFT- and SFT+KL-generated buffers than on DFT- and ASFT-generated buffers. This is consistent with different levels of agreement between the fitting models and the selected targets across data histories.

\FloatBarrier

\subsection{Single-round adaptation to fictional documents}\label{sec:A.9}

\subsubsection*{Study design}

We use FictionalQA (\hyperlink{cite.kirchenbauer2026}{Kirchenbauer et al., 2026}) to study how the four training objectives fit newly supplied documents in a single round of adaptation. We compare SFT, DFT, SFT+KL, and ASFT, together with background-only SFT as a diagnostic control. Each condition starts from the same 4-bit Qwen2.5-Coder-7B-Instruct base and is trained for 700 updates without QA supervision. The four fiction treatments share their document schedule within each training seed. Seeds 42, 43, and 44 provide three training replicates, complementing the variation across starting datasets in the recursive agent study.

The training corpus comprises 990 fiction documents and 512 background documents. In the four fiction treatments, the first 50 updates use only background data. From updates 51 to 700, fiction supplies 5\% of training blocks (8 of every 160). We evaluate checkpoints at updates 50, 175, 350, 525, and 700 on fixed panels of 1,754 exposed-event and 937 held-out-event questions. Answer NLL is the unweighted mean of each answer's token-mean negative log likelihood, in nats/token. Primary four-option MCQ accuracy selects the option with the greatest summed log likelihood. Document NLL pools losses across scored tokens. Retention uses reference-answer NLL on the same 200 TriviaQA items at the base and final checkpoints.

\begin{figure}[H]
\centering
\includegraphics[width=\linewidth,height=.74\textheight,keepaspectratio]{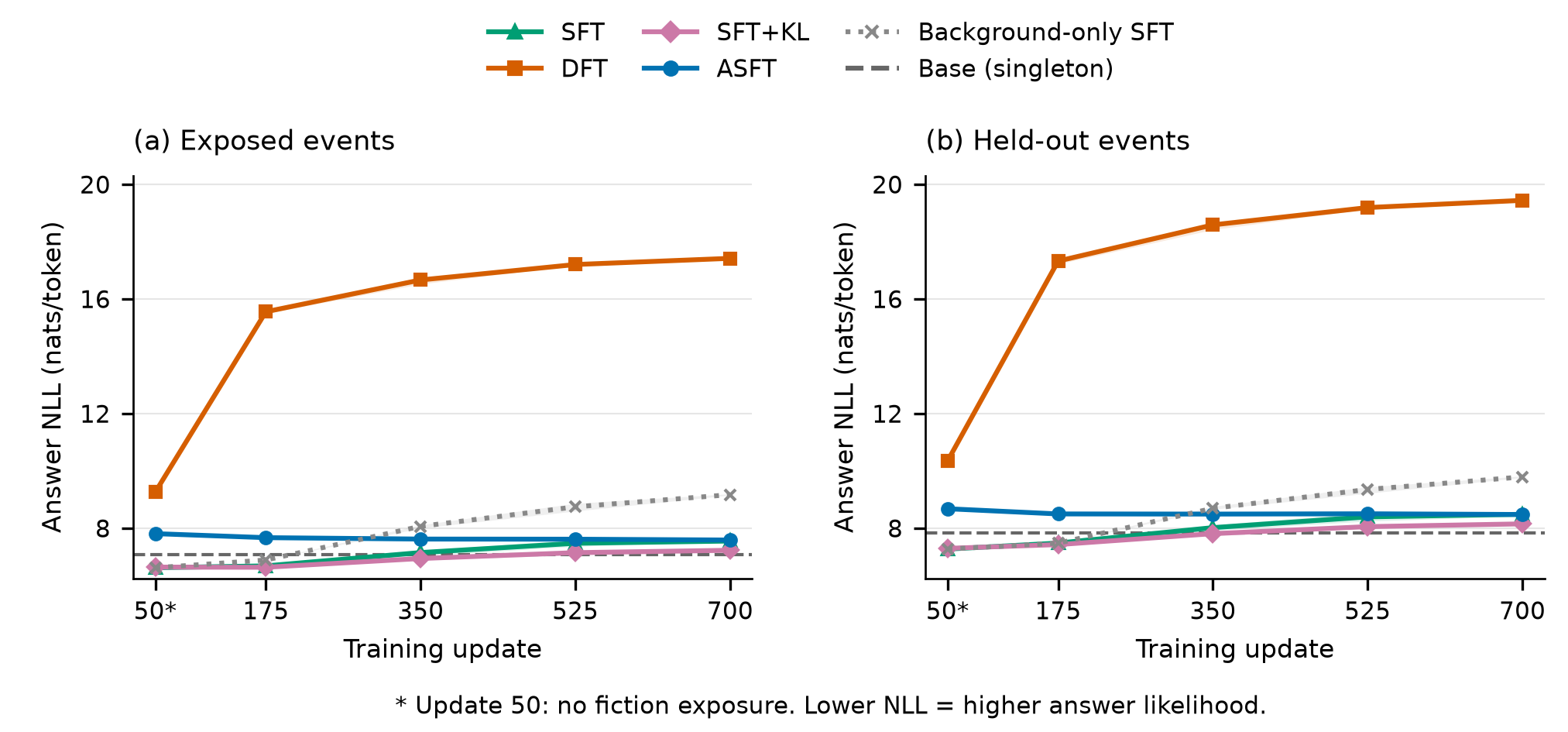}
\caption[SFT+KL achieves the lowest final QA NLL among fitted conditions]{\textbf{SFT+KL achieves the lowest final QA NLL among fitted conditions.} Panels show item-mean answer NLL on (a) 1,754 exposed-event and (b) 937 held-out-event questions. Lower values indicate higher answer likelihood. Lines show three-seed means and shading shows observed ranges; dashed references show the shared base. Update 50 precedes fiction exposure. Every fitted condition finishes above the base on held-out answer NLL.}
\label{fig:A8}
\end{figure}

Means weight the three training seeds equally; ranges report their observed minimum and maximum. The base is one shared evaluation. Exposed and held-out denote fixed event partitions.

\subsubsection*{Document fitting, answer likelihood, and retention}

\begin{figure}[H]
\centering
\includegraphics[width=.92\linewidth,height=.74\textheight,keepaspectratio]{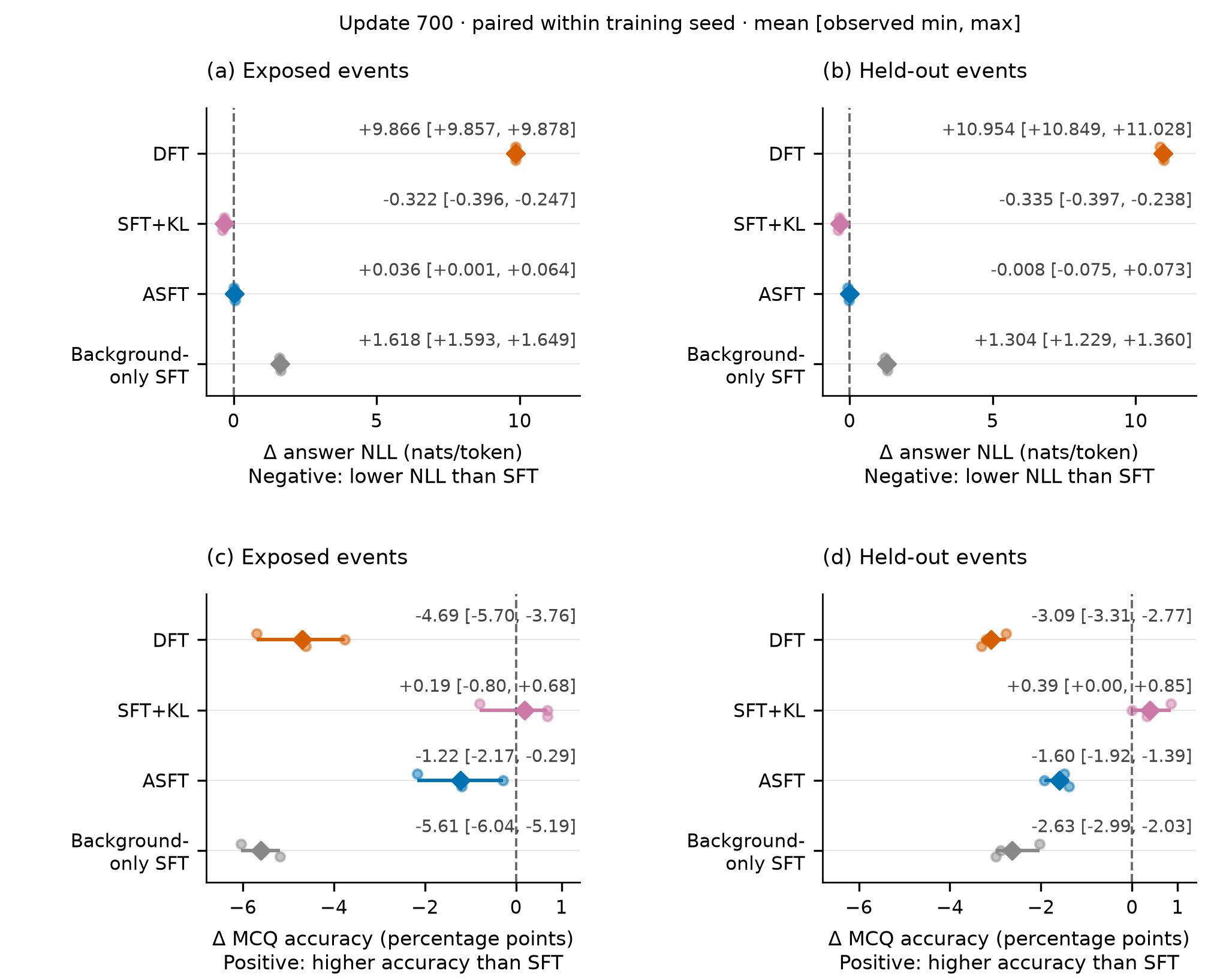}
\caption[SFT+KL improves final answer likelihood relative to SFT across all three seeds]{\textbf{SFT+KL improves final answer likelihood relative to SFT across all three seeds.} Differences are condition minus same-seed SFT at update 700: answer NLL in (a--b), and primary summed-log-likelihood MCQ accuracy in (c--d), for exposed and held-out events. Negative NLL differences and positive accuracy differences favor the condition. Light dots show individual seed differences, dark diamonds their means, and bars their observed minimum--maximum ranges.}
\label{fig:A9}
\end{figure}

SFT and SFT+KL improve fiction-document likelihood relative to the base in both event partitions (\hyperref[fig:A11]{Fig.~A11}). SFT+KL also achieves the lowest final answer NLL among the fitted conditions (\hyperref[fig:A8]{Figs.~A8}--\hyperref[fig:A9]{A9}). On held-out questions, its mean NLL is 8.1492 nats/token, compared with 8.4840 for SFT, 8.4764 for ASFT, and 19.4383 for DFT. Its paired difference from SFT is \ensuremath{-0.3348} nats/token, with an observed range of [\ensuremath{-0.3972}, \ensuremath{-0.2376}] across seeds. Every fitted condition finishes above the base's held-out answer NLL of 7.8207.

\begin{figure}[H]
\centering
\includegraphics[width=\linewidth,height=.74\textheight,keepaspectratio]{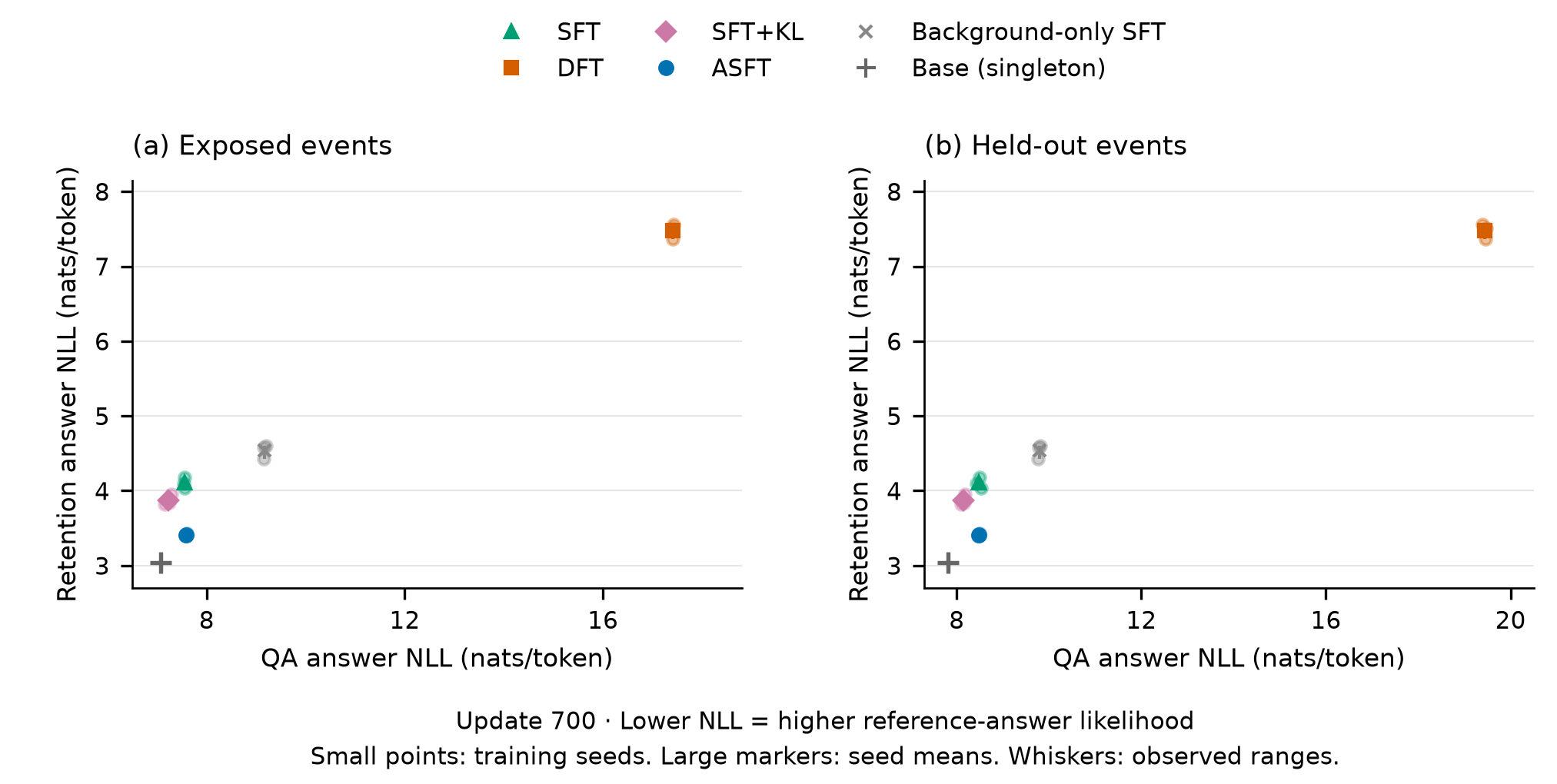}
\caption[ASFT preserves the most TriviaQA likelihood among fitted conditions]{\textbf{ASFT preserves the most TriviaQA likelihood among fitted conditions.} The horizontal axis shows final FictionalQA answer NLL for (a) exposed events and (b) held-out events. The vertical axis shows reference-answer NLL on 200 TriviaQA items. Lower values indicate higher likelihood on both axes. Small points show individual training seeds, larger markers their means, and whiskers marginal observed ranges. The gray plus marks the shared base, which has lower retention and held-out answer NLL than every fitted condition.}
\label{fig:A10}
\end{figure}

\begin{figure}[H]
\centering
\includegraphics[width=.93\linewidth,height=.74\textheight,keepaspectratio]{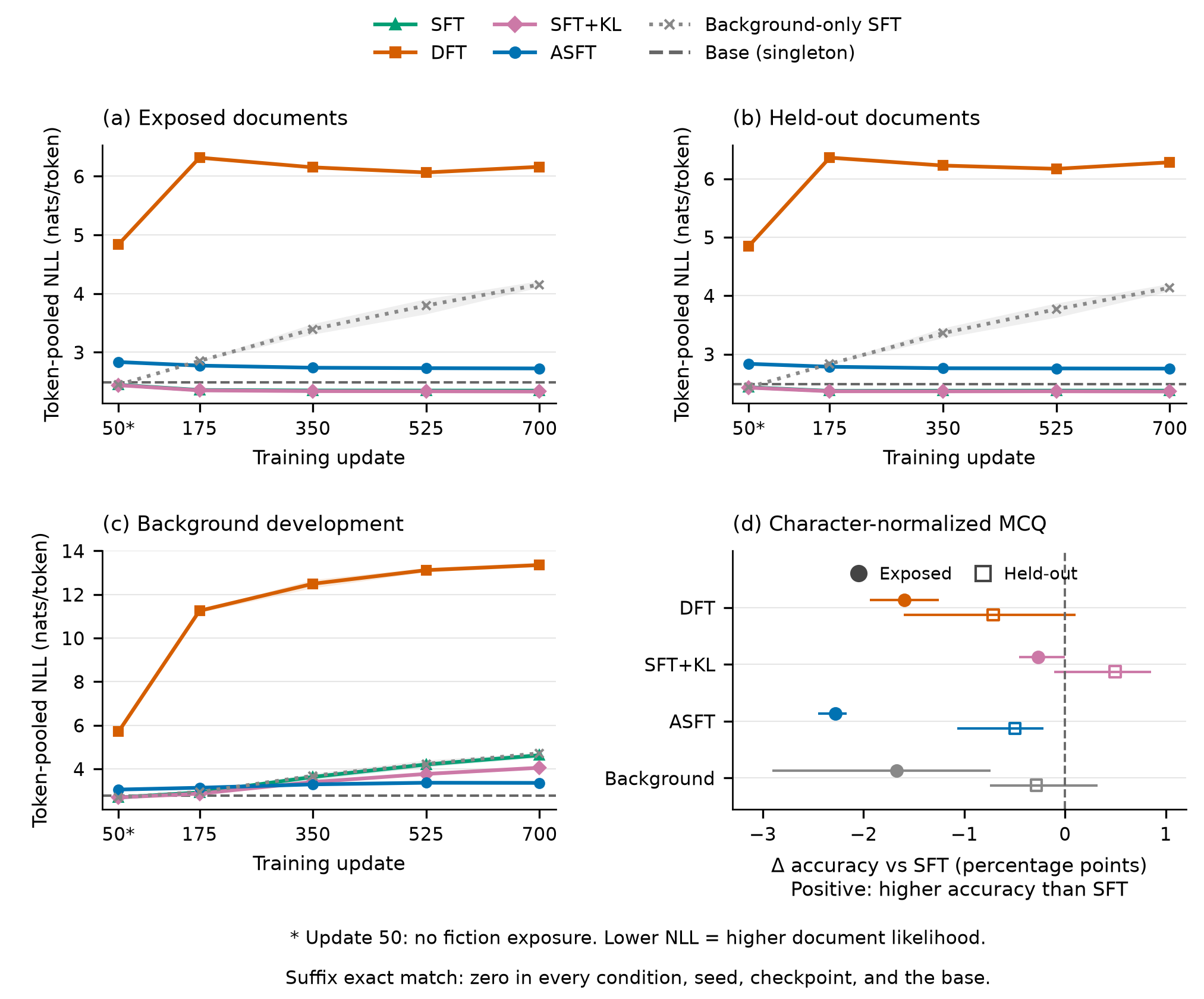}
\caption[SFT and SFT+KL improve fiction-document likelihood in both event partitions]{\textbf{SFT and SFT+KL improve fiction-document likelihood in both event partitions.} Panels (a--c) show token-pooled NLL on 100 exposed fiction documents, 100 held-out fiction documents, and 32 disjoint background-development documents. Lines show seed means, shading observed ranges, and dashed references the shared base. Panel (d) shows update-700 secondary MCQ accuracy differences from same-seed SFT, using summed option log likelihood divided by original choice character length. Filled circles denote exposed events and hollow squares held-out events. Evaluated 50-token suffix exact match is zero for every condition, checkpoint, seed, and the base.}
\label{fig:A11}
\end{figure}

ASFT preserves the most TriviaQA likelihood among the fitted conditions (\hyperref[fig:A10]{Fig.~A10}). Its final retention NLL is 3.4133, compared with 4.1033 for SFT and 3.0422 for the base. Its held-out answer-NLL difference from SFT is small and changes sign across seeds: \ensuremath{-0.0077} [\ensuremath{-0.0753}, 0.0735].

At update 700, SFT scored higher than ASFT on four-option accuracy in all three seeds on both panels (\hyperref[fig:A9]{Fig.~A9c--d}). Mean accuracy on exposed events was 35.46\% for SFT and 34.25\% for ASFT. On held-out events it was 32.69\% and 31.09\%, respectively. SFT+KL had the highest mean accuracy on both panels (35.65\% and 33.08\%). SFT and SFT+KL both exceeded the base on exposed events (34.44\%), while every fitted condition scored below the base on held-out events (35.11\%). Exact recovery of the evaluated 50-token document suffixes is zero throughout. Together, these measurements distinguish improvement in document likelihood, reference-answer likelihood, answer selection, and exact continuation.

\subsubsection*{Evaluation scope}

The study covers one quantized model and fixed evaluation panels. The background corpus repeats for approximately 37 supervised-token epochs in fiction treatments and 39 in the background-only control. At update 700, held-out answer NLL was 9.7878 after background-only SFT, 8.4840 after SFT with fiction, and 7.8207 at the base. Only SFT had a background-only control. Changes from update 50 describe training after the background-only phase.

\FloatBarrier

\subsection{Buffer distances after objective switches}\label{sec:A.10}

\subsubsection*{Distance and comparison design}

Let $A_g$ denote the selected collection generated at model round $g$. The buffer used to fit the model at round $t$ is $B_t=(A_{t-1},A_{t-2},A_{t-3})$. For each of three starting datasets, we compare DFT\ensuremath{\rightarrow}ASFT with its ASFT-throughout reference from the same starting dataset. For the round-3 switch, $B_3$ is the input to the first ASFT fit. By $B_6$, all three inherited collections have been replaced by collections generated under ASFT, and $B_9$ marks a second complete turnover. For the round-6 switch, $B_6$ is the inherited input and $B_9$ marks the first complete turnover. Both switch schedules use the same three ASFT-throughout references. We also compare the three ASFT-throughout lineages with one another from $B_6$ to $B_9$. Each lineage appears in two of those pairs.

For the primary distance, every valid target-token occurrence in a collection receives equal mass. We compute $W_1$ on a fixed tree grouping occurrences by target token, then by target paired with the preceding 8 tokens, 32 tokens, and full context. Successive edge lengths are $1/4$, $1/8$, $127/1024$, and $1/1024$. For each pair of lineages, we compare their newest collections with each other, then their middle and oldest collections in turn. The largest of these three distances is the ordered-buffer distance. \hyperref[tab:A1]{Tables~A1} and \hyperref[tab:A2]{A2} give each pair equal weight when averaging these maxima. In the equal-episode control (EE), each episode with valid targets receives the same mass, distributed uniformly over its targets. The fixed-variant control (FV) assigns each of the five task variants mass $1/5$ and weights episodes equally within each variant.

We also compute $W_1$ after pooling the three collections within each fitting corpus. Each collection then contributes in proportion to its share of valid target tokens. We report this distance both for selected collections alone and after adding the common 500 rehearsal rows at each fit's own token share. In Run 1, over the first turnover after the round-6 switch, the largest distance between collections of matching age fell from 0.645 to 0.620. For the same switched and ASFT-throughout lineages, the distance between the pooled inputs including rehearsal rose from 0.534 to 0.566.

\subsubsection*{Ordered-buffer distances}

\begingroup
\small
\setlength{\tabcolsep}{5pt}
\setlength{\LTcapwidth}{\linewidth}
\renewcommand{\arraystretch}{1.12}
\begin{longtable}{@{}llrrrrr@{}}
\caption{\textbf{Round-3 DFT\ensuremath{\rightarrow}ASFT and matched ASFT-throughout lineages.} Values are ordered-buffer distances at each fit round. Run 1's shared round-0 payload lacks the variant labels needed to compute the fixed-variant distance at round 3. Its round-6 and round-9 collections have complete variant coverage.}\label{tab:A1}\\
\toprule
\textbf{Run} & \textbf{Weighting} & \textbf{Round 3} & \textbf{Round 6} & \textbf{Round 9} & \shortstack[r]{\textbf{Change,}\\\textbf{rounds 3--6}} & \shortstack[r]{\textbf{Change,}\\\textbf{rounds 6--9}} \\
\midrule
\endfirsthead
\multicolumn{7}{@{}l}{\small\itshape Table~\thetable{} (continued)}\\
\toprule
\textbf{Run} & \textbf{Weighting} & \textbf{Round 3} & \textbf{Round 6} & \textbf{Round 9} & \shortstack[r]{\textbf{Change,}\\\textbf{rounds 3--6}} & \shortstack[r]{\textbf{Change,}\\\textbf{rounds 6--9}} \\
\midrule
\endhead
\midrule
\endfoot
\bottomrule
\endlastfoot
Run 1 & Tokens & $0.599622$ & $0.607386$ & $0.634088$ & $+0.007764$ & $+0.026701$ \\*
 & EE & $0.596415$ & $0.603977$ & $0.631113$ & $+0.007561$ & $+0.027136$ \\*
 & FV & --- & $0.604244$ & $0.631487$ & --- & $+0.027243$ \\
\addlinespace[3pt]
Run 2 & Tokens & $0.526621$ & $0.556570$ & $0.584476$ & $+0.029949$ & $+0.027906$ \\*
 & EE & $0.533901$ & $0.554208$ & $0.584986$ & $+0.020306$ & $+0.030779$ \\*
 & FV & $0.533402$ & $0.554349$ & $0.581552$ & $+0.020947$ & $+0.027203$ \\
\addlinespace[3pt]
Run 3 & Tokens & $0.556238$ & $0.606975$ & $0.666045$ & $+0.050737$ & $+0.059070$ \\*
 & EE & $0.552184$ & $0.608820$ & $0.665967$ & $+0.056636$ & $+0.057147$ \\*
 & FV & $0.553914$ & $0.608588$ & $0.666061$ & $+0.054673$ & $+0.057473$ \\
\addlinespace[3pt]
\shortstack[l]{Mean of\\three runs} & Tokens & $0.560827$ & $0.590311$ & $0.628203$ & $+0.029483$ & $+0.037892$ \\*
 & EE & $0.560833$ & $0.589001$ & $0.627355$ & $+0.028168$ & $+0.038354$ \\*
 & FV & --- & $0.589060$ & $0.626367$ & --- & $+0.037307$ \\
\end{longtable}
\endgroup

\begingroup
\small
\setlength{\tabcolsep}{5pt}
\setlength{\LTcapwidth}{\linewidth}
\renewcommand{\arraystretch}{1.12}
\begin{longtable}{@{}>{\raggedright\arraybackslash}p{.38\linewidth}lrrr@{}}
\caption{\textbf{Round-6 DFT\ensuremath{\rightarrow}ASFT pairs and ASFT-throughout pairs.} Switch rows pair each switched lineage with its same-start reference. The other rows pair ASFT-throughout lineages from different starting datasets.}\label{tab:A2}\\
\toprule
\textbf{Pair} & \textbf{Weighting} & \textbf{Round 6} & \textbf{Round 9} & \shortstack[r]{\textbf{Change,}\\\textbf{rounds 6--9}} \\
\midrule
\endfirsthead
\multicolumn{5}{@{}l}{\small\itshape Table~\thetable{} (continued)}\\
\toprule
\textbf{Pair} & \textbf{Weighting} & \textbf{Round 6} & \textbf{Round 9} & \shortstack[r]{\textbf{Change,}\\\textbf{rounds 6--9}} \\
\midrule
\endhead
\midrule
\endfoot
\bottomrule
\endlastfoot
Switch, Run 1 & Tokens & $0.645490$ & $0.620215$ & $-0.025275$ \\*
 & EE & $0.641435$ & $0.634961$ & $-0.006473$ \\*
 & FV & $0.642560$ & $0.633731$ & $-0.008828$ \\
\addlinespace[3pt]
Switch, Run 2 & Tokens & $0.602021$ & $0.635436$ & $+0.033415$ \\*
 & EE & $0.599419$ & $0.630776$ & $+0.031357$ \\*
 & FV & $0.599227$ & $0.630419$ & $+0.031192$ \\
\addlinespace[3pt]
Switch, Run 3 & Tokens & $0.606865$ & $0.672646$ & $+0.065781$ \\*
 & EE & $0.609912$ & $0.669761$ & $+0.059849$ \\*
 & FV & $0.610327$ & $0.669060$ & $+0.058733$ \\
\addlinespace[3pt]
ASFT throughout, Run 1--Run 2 & Tokens & $0.570758$ & $0.603035$ & $+0.032277$ \\*
 & EE & $0.564587$ & $0.603986$ & $+0.039399$ \\*
 & FV & $0.565887$ & $0.602606$ & $+0.036718$ \\
\addlinespace[3pt]
ASFT throughout, Run 1--Run 3 & Tokens & $0.524307$ & $0.599732$ & $+0.075425$ \\*
 & EE & $0.517473$ & $0.597558$ & $+0.080085$ \\*
 & FV & $0.517971$ & $0.598092$ & $+0.080121$ \\
\addlinespace[3pt]
ASFT throughout, Run 2--Run 3 & Tokens & $0.594921$ & $0.634003$ & $+0.039083$ \\*
 & EE & $0.590826$ & $0.623630$ & $+0.032805$ \\*
 & FV & $0.592724$ & $0.627183$ & $+0.034459$ \\
\addlinespace[3pt]
Mean switch pair & Tokens & $0.618125$ & $0.642766$ & $+0.024641$ \\*
 & EE & $0.616922$ & $0.645166$ & $+0.028244$ \\*
 & FV & $0.617371$ & $0.644403$ & $+0.027032$ \\
\addlinespace[3pt]
Mean ASFT-throughout pair & Tokens & $0.563329$ & $0.612257$ & $+0.048928$ \\*
 & EE & $0.557628$ & $0.608391$ & $+0.050763$ \\*
 & FV & $0.558861$ & $0.609294$ & $+0.050433$ \\
\end{longtable}
\endgroup

\subsubsection*{Pooled fitting inputs and token shares}

At round 6 in Run 1, the newest, middle, and oldest selected collections accounted for 34.4\%, 36.7\%, and 22.5\% of valid target tokens in the switched input. Rehearsal accounted for the remaining 6.4\%. The corresponding shares in the ASFT-throughout input were 30.8\%, 31.9\%, 31.0\%, and 6.3\%. We reconstructed the Run 1 ASFT-throughout corpora for rounds 6 and 9 with the historical encoder. The training recipe specified one pass through each corpus, but row-level records of optimizer exposure were unavailable for those two fits.

\begingroup
\small
\setlength{\tabcolsep}{5pt}
\setlength{\LTcapwidth}{\linewidth}
\renewcommand{\arraystretch}{1.12}
\begin{longtable}{@{}llrrr@{}}
\caption{\textbf{Mean Wasserstein distance between pooled fitting inputs.} Each mean averages the three corresponding pairwise distances. For the round-3 switch, the pooled distance including rehearsal increased from rounds 6 to 9 by 0.0342 in Run 1, 0.0298 in Run 2, and 0.0683 in Run 3.}\label{tab:A3}\\
\toprule
\textbf{Comparison} & \textbf{Pooled corpus} & \textbf{Round 3} & \textbf{Round 6} & \textbf{Round 9} \\
\midrule
\endfirsthead
\multicolumn{5}{@{}l}{\small\itshape Table~\thetable{} (continued)}\\
\toprule
\textbf{Comparison} & \textbf{Pooled corpus} & \textbf{Round 3} & \textbf{Round 6} & \textbf{Round 9} \\
\midrule
\endhead
\midrule
\endfoot
\bottomrule
\endlastfoot
Round-3 switch & Selected only & $0.361548$ & $0.567118$ & $0.614570$ \\*
 & Selected + rehearsal & $0.339071$ & $0.532119$ & $0.576225$ \\
\addlinespace[3pt]
Round-6 switch & Selected only & --- & $0.571466$ & $0.615685$ \\*
 & Selected + rehearsal & --- & $0.539942$ & $0.578933$ \\
\addlinespace[3pt]
ASFT throughout & Selected only & --- & $0.522859$ & $0.603262$ \\*
 & Selected + rehearsal & --- & $0.490307$ & $0.563991$ \\
\end{longtable}
\endgroup

\Needspace{.66\textheight}

\subsection{Experimental parameters}\label{sec:A.11}

Both studies fit fresh LoRA adapters on the 4-bit \texttt{unsloth/\allowbreak{}Qwen2.5-\allowbreak{}Coder-\allowbreak{}7B-\allowbreak{}Instruct-\allowbreak{}bnb-\allowbreak{}4bit} base. We used LoRA rank 32, scaling parameter $\alpha=32$, and zero dropout. Both studies used paged 8-bit AdamW, zero weight decay, a maximum gradient norm of 1.0, a per-device microbatch size of 1, and 32 gradient-accumulation steps. SFT+KL and ASFT used a forward-KL coefficient of 0.1 with the frozen base as reference (\hyperref[sec:3.2]{\S{}3.2}).

\begingroup
\small
\setlength{\tabcolsep}{5pt}
\renewcommand{\arraystretch}{1.15}
\begin{longtable}{@{}>{\raggedright\arraybackslash}p{.203\linewidth}>{\raggedright\arraybackslash}p{.347\linewidth}>{\raggedright\arraybackslash}p{\dimexpr.45\linewidth-4\tabcolsep\relax}@{}}
\toprule
\textbf{Parameter} & \textbf{Recursive file cleanup} & \textbf{Single-round FictionalQA} \\
\midrule
\endfirsthead
\toprule
\textbf{Parameter} & \textbf{Recursive file cleanup} & \textbf{Single-round FictionalQA} \\
\midrule
\endhead
\midrule
\endfoot
\bottomrule
\endlastfoot
Training budget & One epoch per fit, for up to 10 fitted rounds & 700 optimizer updates \\
\addlinespace[3pt]
Maximum training sequence length & 16,384 tokens & 2,048 tokens \\
\addlinespace[3pt]
Learning-rate schedule & Peak $2\times10^{-4}$, 3\% warmup, cosine decay & 50 warmup updates to $5\times10^{-5}$, then cosine decay to $5\times10^{-6}$ \\
\addlinespace[3pt]
Training data & Up to three recent selected collections and 500 fixed rehearsal rows from \texttt{flytech/\allowbreak{}python-\allowbreak{}codes-\allowbreak{}25k} per fit & 990 fiction and 512 background documents. The first 50 updates used background only. Thereafter, fiction supplied 8 of every 160 blocks in the four fiction conditions. \\
\addlinespace[3pt]
Supervised tokens & Completion tokens in selected and rehearsal examples & Raw document tokens, with the first token of each block and padding masked \\
\addlinespace[3pt]
Replication & Three starting collections, each shared across objectives. Adapter seed 42 and data-order seed 3407. & Three training seeds, 42, 43, and 44. The four fiction conditions shared a document schedule within each seed. \\
\addlinespace[3pt]
Evaluation & The same 200 items and inference seeds at each checkpoint, with 40 items per filesystem variation & Fixed panels at updates 50, 175, 350, 525, and 700 (\hyperref[sec:A.9]{\S{}A.9}) \\
\end{longtable}
\endgroup

Each cleanup collection targeted 1,000 selected rows, with up to 12 rows per successful episode. Success required removing at least 5\% of disposable files, preserving protected files, and leaving the system usable. A lineage stopped if it failed to reach the collection target within 4,000 attempts.

\clearpage

\begingroup
\makeatletter
\renewcommand{\@biblabel}[1]{}
\makeatother
\setlength{\labelsep}{1.2em}

\endgroup

\end{document}